\documentclass[preprint,12pt]{elsarticle}

\usepackage{amssymb}

\usepackage{amsmath,amsfonts}
\usepackage{algorithmic}
\usepackage{algorithm}
\usepackage{array}
\usepackage{textcomp}
\usepackage{stfloats}
\usepackage{url}
\usepackage{verbatim}
\usepackage{graphicx}
\usepackage{caption}
\usepackage{hyperref}
\usepackage{enumitem}
\usepackage{subfigure}
\usepackage{ragged2e}
\usepackage{booktabs}
\usepackage{multirow}
\usepackage{soul}
\usepackage{xcolor}

\sethlcolor{yellow} 

\journal{Robotics and Autonomous Systems}

\begin{document}

\begin{frontmatter}



\title{Early Detection of Human Handover Intentions in Human-Robot Collaboration: \\ Comparing EEG, Gaze, and Hand Motion} 


\author[label1]{Parag Khanna \fnref{label3,label4}}
\author[label1]{Nona Rajabi \fnref{label3}}
\author[label2]{Sumeyra U. Demir Kanik}
\author[label1]{Danica Kragic}
\author[label1]{Mårten Björkman}
\author[label1]{Christian Smith}
\fntext[label3]{These authors contributed equally to this work}
\fntext[label4]{To whom correspondence should be addressed; E-mail: paragk@kth.se.}

\affiliation[label1]{organization={KTH Royal Institute of Technology, Stockholm},
             city={Stockholm},
             country={Sweden}}

\affiliation[label2]{organization={Ericsson Research},
             city={Stockholm},
             country={Sweden}}

\begin{abstract}
Human-robot collaboration (HRC) relies on accurate and timely recognition of human intentions to ensure seamless interactions. 
Among common HRC tasks, human-to-robot object handovers have been studied extensively for planning the robot's actions during object reception, assuming the human intention for object handover. However, distinguishing handover intentions from other actions has received limited attention. Most research on handovers has focused on visually detecting motion trajectories, which often results in delays or false detections when trajectories overlap. 
This paper investigates whether human intentions for object handovers are reflected in non-movement-based physiological signals. We conduct a multimodal analysis comparing three data modalities: electroencephalogram (EEG), gaze, and hand-motion signals. Our study aims to distinguish between handover-intended human motions and non-handover motions in an HRC setting, evaluating each modality's performance in predicting and classifying these actions before and after human movement initiation.
We develop and evaluate human intention detectors based on these modalities, comparing their accuracy and timing in identifying handover intentions. To the best of our knowledge, this is the first study to systematically develop and test intention detectors across multiple modalities within the same experimental context of human-robot handovers. 
Our analysis reveals that handover intention can be detected from all three modalities. Nevertheless, gaze signals are the earliest as well as the most accurate to classify the motion as intended for handover or non-handover.
\end{abstract}

\end{frontmatter}




\section{Introduction}
Human-Robot Collaboration (HRC) is a critical component of the Industry 4.0 revolution, boosting productivity and efficiency in various industries, including manufacturing and logistics 
\cite{HRC_industry_4,survey_review_2022_object_handovers}. The transition from separate workspaces to shared environments where humans and robots collaborate allows for a more integrated approach to task execution. In this collaborative setting, humans and robots share responsibilities, each bringing their own strengths to improve overall operational effectiveness \citet{human_role_safety_collab_industry4}. 
Thus, understanding human intent and facilitating seamless interactions between humans and robots is essential for maximizing the benefits of HRC \cite{comm_intent_before_HRC_strabala2012,intention_multimodal_Wang2022}. A critical aspect of these interactions is 
recognizing human intent as early as possible and ideally, before actions occur \cite{non_invasive_real_time_activity_track_human,gaze_vs_hand_moving_target,comm_intent_before_HRC_strabala2012}. 
Traditional approaches often rely on visual cues and hand-tracking to detect intentions, but these methods may not always provide timely or accurate predictions, especially in complex collaborative environments \cite{intention_multimodal_Wang2022, comm_intent_before_HRC_strabala2012} where similar initial actions may lead to different outcomes. 
This work focuses on the use of non-invasive, non-movement-based physiological signals to recognize human intent in human-robot interactions, using handovers as a case study\footnote[3]{All codes for data processing and model training presented in this study are published in the following URL: {\small \url{https://github.com/NonaRjb/Human-to-Robot-Handover-Detection.git}}. Also, the data can be accessed from the following URL and has been made publicly available: \small{\url{https://doi.org/10.5281/zenodo.14876712}}}. Signals like EEG and gaze can provide insights into human intentions and motions without relying on physical movement. As such, they offer a potential alternative to movement-dependent, vision-based methods for monitoring human actions, warranting evaluation and comparison against conventional approaches.

Handover, the act of passing an object to another person, is a crucial part of daily human interactions~\cite{survey_review_2022_object_handovers,When_where_how_human-human_studyStrabala}. Therefore, mastering handover skills is essential for robots to function effectively as collaborative partners \cite{survey_review_2022_object_handovers,data_driven_grip_release}. It is vital for a robot to recognize the human's intention to hand over an object and initiate the receiving action at the right time to avoid delays or failures \cite{intention_multimodal_Wang2022,nonaEEG}. Such failures can erode human trust and reduce the willingness to collaborate with robots in the future. Methods based on visual sensing for hand tracking can generate robot motion only after the human initiates movement for handover.
In contrast, approaches using physiological signals such as gaze and EEG may allow earlier intention detection, as they do not depend directly on physical movement. Human gaze behaviors during handovers have been studied to inspire robotic gaze in human-robot handovers \cite{shared_attention_gaze_hanover_timing_2014}. Meanwhile, EEG signals, which measure brain activity non-invasively, have shown potential in brain-computer interfaces (BCIs) to interpret human intentions in various contexts, including human-robot interaction \cite{varbu2022past}.
We develop and evaluate human intention detectors based on three different data modalities—EEG, gaze, and hand-tracking (by visual sensing)—comparing their accuracy and timing in detecting the intention to hand over an object to a robot. To the best of our knowledge, this is the first study to systematically develop and test intention detectors across these diverse modalities within the exact same experimental context of human-robot handovers. 

We investigate whether EEG or gaze can enhance intention detection performance beyond the hand-tracking baseline, enabling earlier and more accurate detection of handover intentions in close-proximity human-robot collaboration. The novelty of our study lies in: (i) conducting an experimental study to recognize human-to-robot handover intentions while isolating the intention from confounding factors such as movement direction and sequence; (ii) proposing and demonstrating EEG- and gaze-based intention detectors for human-robot handovers; and (iii) directly comparing these methods to traditional hand-tracking approaches, as used in \cite{contrast_handover_poses_for_intent_handover_cakmak2011,survey_review_2022_object_handovers}.

Our research questions are:
\begin{enumerate}
    \item RQ1: To what extent can EEG or gaze be used to classify the upcoming motion for handover or non-handover before the motion has started? 
    \item RQ2: How early can a human intention to hand an object to a robot be detected based on different modalities?
\end{enumerate}
In addition to our main goals, we assess how different classifiers affect intention detection performance. Specifically, we compare a classic linear classifier with a neural network for each modality. Given the ability of neural networks to identify complex patterns without extensive dimensionality reduction \cite{wang2016action,rudenko2020human,craik2019deep,tang2017single,george2016real}, we investigate whether they outperform linear models in detecting task-related patterns. In summary, our main contributions are:
\begin{itemize}
    \item We compare EEG, gaze, and hand motion trajectory modalities for their accuracy and timing in predicting whether upcoming human motion is an object handover.
    \item We demonstrate that gaze data is the fastest and most accurate in distinguishing handover intentions, showing predictive power even during the planning phase before movement begins.
    \item We show that combining these modalities in a multimodal model improves the performance of the weaker modality both before and after movement initiation.
    \item We introduce a hybrid dataset with simultaneous EEG, gaze, and hand motion trajectory recordings from 15 participants performing handover and non-handover motions in a human-robot collaboration setting.
\end{itemize}

\section{Related Work}
Recent advances in robotics have enabled robots to integrate into human environments, working alongside people in various settings~\cite{survey_review_2022_object_handovers}. Since handover is a crucial part of human interactions, human-robot handovers have been extensively studied in both human-human \cite{human_human_to_human_robot_study_Controzzi,datasetmultimodalhandovers} and human-robot interactions \cite{survey_review_2022_object_handovers,human_human_to_human_robot_study_Controzzi,When_where_how_human-human_studyStrabala}. A handover consists of two phases: the \emph{pre-handover} phase, which includes communicating the intent, and the \emph{physical exchange} of the object~\cite{survey_review_2022_object_handovers}. This work focuses on recognizing the human giver's intention in human-to-robot handovers. In particular, we compare different data modalities on their accuracy and timing in detecting human intention to hand over an object to a robot.

Common signals indicating the intention to hand over an object include moving closer to the receiver \cite{human_dog_behaviors_in_handovers} and using communication cues such as gaze \cite{shared_attention_gaze_hanover_timing_2014}, body posture, and verbal signals \cite{survey_review_2022_object_handovers}. Additional cues include reaching movements, leaning towards the receiver, the giver's orientation, hand occupancy, and gaze direction \cite{comm_intent_before_HRC_strabala2012}. Robotic vision is often used to predict human-giver movements during handovers, aiding in the planning and execution of the robot-taker's arm motion. However, this approach generally assumes the intention to hand over an object rather than explicitly detecting it \cite{survey_review_2022_object_handovers}. In \cite{basic_CV_for_intention_kwan2020handover}, handovers are distinguished from non-handover actions using object detection methods, skeletal key points, and head pose estimation based on gestures like extending an arm with an object in hand. Nevertheless, the features used might not be distinctive in scenarios where the movement sequences are similar for handover and non-handover tasks. Despite this, challenges such as complex backgrounds and constrained working spaces remain significant \cite{intention_multimodal_Wang2022}. A different approach in \cite{handover_with_wearable_EMG_Wang2019} uses a forearm-mounted wearable sensor that relies on electromyography (EMG) signals and arm rotation to classify human handover intentions, categorizing different gestures into actions. However, the considered human ``Give" gesture does not encompass all natural handover scenarios. Another method, the teaching-learning-prediction (TLP) model in \cite{intention_multimodal_Wang2022}, integrates speech, EMG-IMU (Inertial Measurement Unit) arm sensors, and image data to predict human actions during handovers. This approach typically recognizes human intention 1-2 seconds after the movement begins.

Human gaze behaviors during handovers have been studied for various aspects of human-robot handovers. For instance, \cite{shared_attention_gaze_hanover_timing_2014} found that givers in human handovers looked at both the handover location and the receiver’s face. During the handover motion, both givers and receivers are typically focused on each other’s hands \cite{robot_gaze_behav_in_h2r_handovers_kshirsagar}. Research has shown that human reaching movements are often preceded by saccadic eye motions toward the target \cite{gaze_vs_hand_moving_target}. Furthermore, \cite{eye_hand_coordination_2001} found that gaze fixation usually precedes or guides hand movements, indicating that gaze helps in planning hand movements by directing attention to key positions for the hands and objects.

Motion tracking with visual sensing detects human movement, including handovers, only after it begins. If initial hand movements resemble non-handover motions, it can be difficult to determine the intended action and whether the robot should respond, even after the motion starts.
EEG, a non-invasive method for measuring brain activity, is widely used in brain-computer interfaces (BCIs) to enable direct robot control via brain signals \cite{varbu2022past,lyu22, mondini2020continuous}. Previous research has successfully differentiated EEG signals associated with motor preparation from resting states \cite{MAMMONE2020source, 9534028}. However, identifying the type of the action is challenging, especially when the movement directions and body parts are identical across different actions. As a result, most studies on classifying movement intentions focus on significantly different trajectories~\cite{MAMMONE2020source, gordienko21, MOHSENI2020, ofner2019}. Recently, \cite{cooper20} showed that EEG signals differ between handover intentions and solo object movements, but their analysis was limited to group-level statistics, comparing average signals across conditions. Furthermore, since they only examined two scenarios—solo object movement and object handover—it remains unclear whether the observed differences stem from the presence of a robot or are truly indicative of the handover task. In our previous work~\cite{nonaEEG}, we addressed this limitation by introducing a third condition in which the human collaborates with the robot without a direct object handover. Our results revealed distinct EEG signals for handover intentions compared to other actions, even when non-direct joint actions were included. Furthermore, we demonstrated that single trials from different conditions could be successfully classified, although with lower accuracy compared to detecting general movement intentions versus no-movement conditions. These findings indicate the feasibility of using EEG signals to detect human intentions in object handovers to robots.

While the prior research has explored the use of gaze tracking, EEG signals, and motion tracking to detect human handover intentions, two key gaps remain. 

First, to the best of our knowledge, an intention detector utilizing these modalities to classify handover or non-handover movements has not been developed or implemented in real-world scenarios.

Second, there is no direct comparison of the efficacy of these different modalities for detecting handover intentions in the same experiment. These gaps highlight the need for a comprehensive approach that not only builds a system using multiple modalities but also carefully tests and compares their effectiveness.
In this work, we compare \emph{EEG}, \emph{gaze}, and \emph{hand-tracking} (via visual sensing) to detect and predict human handover intentions in close-proximity human-robot collaboration. We analyze whether handover intentions differ from other types of intentions across these three modalities, particularly when movement direction and sequences are controlled for similarity. Unlike many prior studies, we do not assume handover intention in advance. Rather than focusing on extracting sophisticated features, we explore how effectively each modality distinguishes handover intentions from non-handover intentions.

\section{Experimental Design}
\begin{figure}[t]
    \centering
    \includegraphics[width=\linewidth]{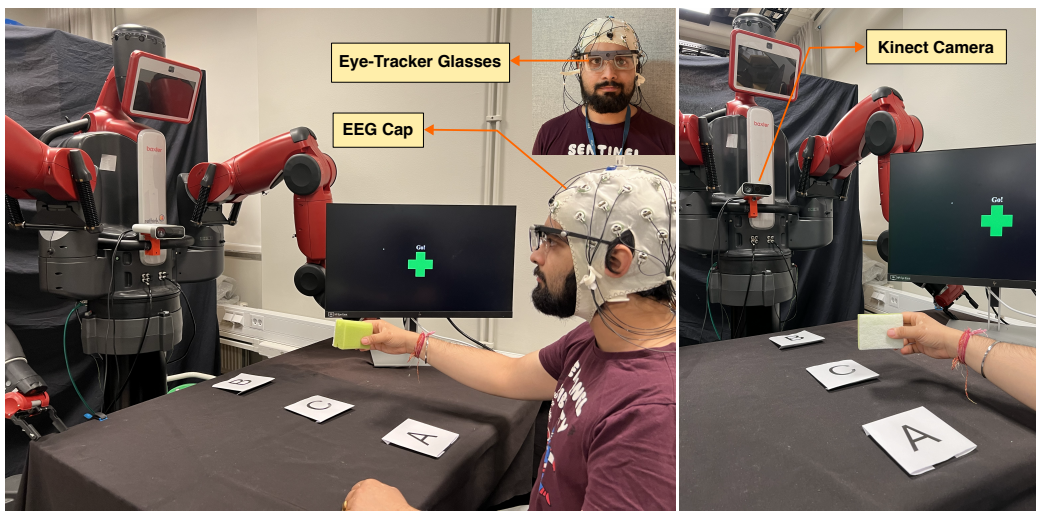}
    \caption{\justifying Experimental Setup: The participant sits across a table from the Baxter robot, wearing the EEG cap and Tobii eye tracker glasses. An Azure Kinect RGBD camera is installed on the robot's torso.}
    \label{fig:experimental setup}
\end{figure}

We designed a within-subject experiment involving human participants interacting with a Baxter robot~\cite{baxter-Doe:2024:Online}. Each participant sat across from the robot, wearing an EEG cap and Tobii Pro Glasses 2 eye-tracker glasses. An RGBD camera, Microsoft Azure,  mounted on the robot's torso, tracked the participant's hand motion trajectory. The experimental setup is depicted in Fig.~\ref{fig:experimental setup}. From the participant's point of view, there are three marked positions on the table: A, B, and C in a straight line. In each trial, a green sponge is moved from its initial position A, near the participant, to the final position B, near the robot. The movement is executed through one of three types of human actions (also visualized in Supplementary Fig. 1):
\begin{enumerate}[label=(\alph*)]
    \item \textbf{Solo Action:} This action is performed solely by the participant. \textit{The participant picks the object from A} and places it at B.
    \item \textbf{Handover - Human Giver:} This action involves a human-to-robot handover. \textit{The participant picks the object from A} and hands it to the robot over C. The robot then places the object at B. 
    \item \textbf{Joint Action:} This action is a non-handover collaborative action between the participant and the robot. \textit{The participant picks the object from A} and places it at C. Then, the robot picks the object from C and places it at B. 
\end{enumerate}
\begin{figure}[h]
    \centering
    \includegraphics[width=0.9\linewidth]{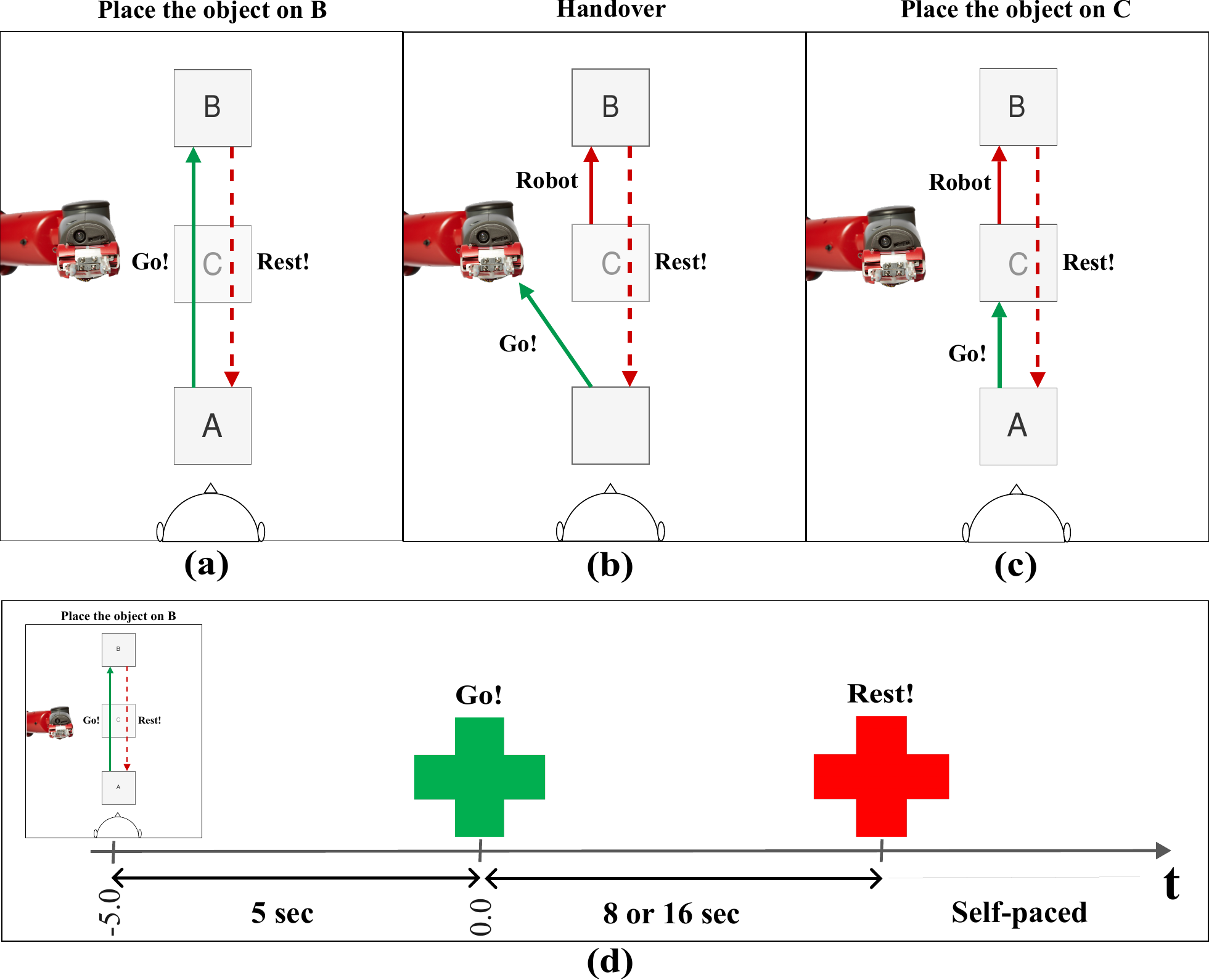}
    \caption{Experimental conditions and timing diagram, Adapted from Fig. 2 in \cite{nonaEEG}. Visual instructions for the three task conditions: \textbf{(a)} solo, \textbf{(b)} handover, and \textbf{(c)} joint actions. \textbf{(d)} Timing diagram of the experiment with $t=0$ at the \emph{Go!} signal.}
    \label{fig:gui-ins}
\end{figure}

\noindent The initial actions, shown in italics, are identical across all three tasks. Further, in both the \emph{Handover} and \emph{Joint} actions, the robot arm was controlled by the experimenter, though participants were unaware of this. 

In our design, joint action specifically refers to a collaborative action with the robot that is not a direct handover.
Each trial corresponds to one of three actions conveyed to the participant via instructions on a screen placed on the table (Fig.~\ref{fig:experimental setup}). There were 90 trials in total, with an equal number, i.e. 30, of each task condition performed in a random order. Fig.~\ref{fig:gui-ins} illustrates the timeline of instructions for each trial. The first image is a cue specifying the action type, displayed for 5 seconds with a countdown from 5 to 1 second. The next image shows a green cross and the word "Go!", signaling the participant to begin the action. For solo actions, this image is displayed for 8 seconds, while for handover and joint actions, it is displayed for 16 seconds to allow sufficient time for task completion based on pilot trials. The final image is a red cross with the word "Rest!", indicating rest time. During this period, the participant returns the object from B to A and presses the space bar on a keyboard to mark the end of the trial and the start of the next one.

\section{Dataset Preparation}
\subsection*{Participants}
A total of 15 individuals (age = 26.67 \(\pm\) 4.25, 11 males, 4 females) participated in the experiment. Six participants had prior experience with robots, mostly from participation in other studies, and all of them were right-handed. All participants were provided with a detailed explanation of the experiment and informed consent was obtained from them before the experiment. The study was approved by the Swedish Ethical Review Authority, Etikprövningsmyndigheten (DNR:2022-01555-01).
Two participants (Subject ID = 7 and 13, both male) were excluded from the EEG analysis due to technical issues with the EEG recording device during the experiment. Two participants (Subject ID = 13 and 14) also had technical issues with hand-tracking; hence, they were excluded from the motion analysis. Thus, participant 13 was excluded from the overall analysis. For all other participants, we include the gaze analysis, although, for a few participants, less trials than the total number of trials (i.e., 90) were available. Please refer to Supplementary Table S3 for the total number of trials per participant per modality.

\subsection*{Data Acquisition and Preprocessing}

We simultaneously recorded EEG, gaze, and hand motion trajectory data while participants performed the experiment tasks. Fig.~\ref{fig:experimental setup} illustrates the experimental setup, including the EEG recording cap, eye-tracking glasses, and the RGBD camera. 

\subsubsection*{EEG Data}

EEG signals were recorded using 32 Ag/AgCl BrainProducts active electrodes~\cite{brainproductsBrainProducts}, positioned at equidistant locations according to the 10–20 system. All EEG preprocessing was performed in Python (v3.9) using the MNE-Python package~\cite{gramfort2013meg}. To eliminate artifacts from muscle activity and blinking, the signal was first band-pass filtered between 1 and 100 Hz and then processed with the MNE-ICALabel package to identify non-brain components using independent component analysis (ICA)~\cite{makeig1995independent,iriarte2003independent}. After removing these components, the signal was low-pass filtered at 40 Hz. Depending on the required data segment, the data was epoched from 5 seconds before to a maximum of 6 seconds after movement onset. Finally, the data was downsampled from 1000 Hz by a factor of 4 to reduce noise as well as computational costs. 

We used only 12 out of the 32 channels recorded by the EEG device, focusing on central and frontal channels, which contain more information related to motor planning and execution~\cite{hanakawa2003functional,chaisaen2020decoding,yu2022effects}. This channel selection reduces data dimensionality and filters out irrelevant brain activity. Specifically, we used the Cz, C3, C4, FC1, FC2, FC5, FC6, CP1, CP2, F3, F4, and Fz channels, as shown in Supplementary Fig. 2.

\subsubsection*{Gaze Data}
Gaze data was acquired using Tobii Glasses 2 \cite{Tobiiglasses:2024:Online} worn by the participants. These glasses provide recorded videos and gaze data. We utilized GlassesViewer \cite{glassviewer}, an open-source software, to extract, filter, and parse the gaze data from the Tobii Glasses 2 recordings. The gaze data is a stream of $(X,Y)$ values (denoted as $GazeX$ and $GazeY$) referenced to the corner $(0,0)$ of the recorded video frame, which has a size of $(1920,1080)$. Linear interpolation was used to handle missing gaze samples in the data stream, a common technique used in similar studies to fill in gaps caused by blinks, head movements, or other sources of noise \cite{glassviewer}. For all participants, the gaze data was referenced to a specific point on the robot's torso (Supplementary Fig. 3). 
This reference point was tracked in the recorded video frames using normalized cross-correlation in MATLAB. The sampling rate for the recorded video is 25 Hz and for each video frame, we track this reference point.
Due to this processing, the final sampling frequency for the gaze data was 25 Hz.

\subsubsection*{Hand Motion Trajectory Data}
The human motion was tracked using the Microsoft Azure Kinect RGBD camera \cite{Azure_Kinect} and a ROS-based implementation of the Azure Kinect Body Tracking SDK. The output of this body-tracking SDK is the tracked human skeleton referenced by the Kinect camera. Since all participants were right-handed, we extracted the right-hand tracking data from the recorded tracking skeleton data for the participants, in particular the X, Y, and Z coordinates. Given the output frequency of 5 Hz for this tracking SDK, the hand-motion data had a final sampling frequency of 5 Hz.

\subsection*{Dataset}

We structure the data from different modalities as time series data to detect whether the upcoming or ongoing motion is a handover or not. This sequential arrangement allows a classification model to capture the temporal dependencies necessary for accurate classification. A time series \(\mathbf{X_T}\) is represented as:

\begin{equation}
    \mathbf{X_T} = \{x_1, x_2, x_3, \ldots, x_T\} ,
\end{equation}

\noindent where \(x_t\) represents the observations at time \(t\), and \(T\) is the total number of time samples. Maintaining the order of these observations is crucial for the model to learn patterns and predict class labels accurately. In our dataset, depending on the modality, \(x_t\) is defined as:

\begin{equation}
    x_t = 
    \begin{cases} 
    \text{Gaze}: [\text{GazeX}_t, \text{GazeY}_t] \\
    \text{Motion}: [\text{HandX}_t, \text{HandY}_t, \text{HandZ}_t] \\
    \text{EEG}: [\text{f}_1, \text{f}_2, ..., \text{f}_F]
    \end{cases},
\end{equation}

\noindent where \(\text{GazeX}_t\) and \(\text{GazeY}_t\) are the 2D gaze coordinates, and \(\text{HandX}_t\), \(\text{HandY}_t\), and \(\text{HandZ}_t\) are the 3D hand coordinates. For EEG signals, we used the time-frequency transformation of the data, with \(\text{f}_1\) to \(\text{f}_F\) indicating the power of frequency components at time \(t\). The use of time-frequency features was based on previous studies that show their effectiveness in predicting motor-related EEG activity~\cite{alazrai2017eeg, xu2018wavelet, MAMMONE2020source}. For more details on the data features, please refer to the \emph{Feature Extraction} in the \emph{Data Analysis} section.

\section{Data Analysis}
In this section, we statistically assess whether the object handover condition differs significantly from the two other conditions in gaze and EEG modalities, focusing on the period before movement initiation. Hand motion trajectories are excluded from this analysis since, by design, no hand movement occurs before the \emph{Go!} signal.

\subsection*{Gaze Data Analysis}
As indicated by previous research, human-givers look at the handover location and the target location for object placement in object manipulation. Thus, we analyze the gaze behavior of the participants in our study by segmenting the regions viewed by participants into four zones: \emph{robot}, \emph{position B}, \emph{position C}, and \emph{other}. The ``robot" zone includes the robot's arm and torso, while the ``other" primarily includes the instruction screen. Table~\ref{tab:gaze_a}(a) and ~\ref{tab:gaze_b}(b) show a trend that differentiates the three human actions in terms of these zones for a particular participant (S8).
\begin{table}[h!]
    \centering
    \caption{Frequency (in \%) of gaze locations falling in one of four different regions denoted as Robot, Position B, Position C, and Other for different experiment conditions. \textbf{(a, c)} present values for times $t=-5$ to $t=0$ (presentation of \emph{Go!} signal and movement onset). \textbf{(b, d)} present values for times $t=-5$ to $t=3$ sec including both before and after movement onset. \textbf{(a, b)} correspond to a representative participant (S8) and \textbf{(c, d)} show median values across all participants. The maximum values (excluding Other) per condition are made bold.}
    \label{tab:participant8}
    
        \begin{minipage}{0.4\textwidth}
            \centering
            \resizebox{\textwidth}{!}{
            \begin{tabular}{cccc}
                \toprule
                & \textit{Solo} & \textit{Handover} & Joint \\
                \midrule
                Robot & 11.85 & 15.28 & 2.10\\
                \midrule
                Pos. B & 17.37 & 4.25 & 2.39\\
                \midrule
                Pos. C & 2.29 & 6.40 & 18.74\\
                \midrule
                Other & 68.50 & 74.07 & 76.77\\
                \bottomrule
            \end{tabular}}
            \label{tab:gaze_a}
            \textbf{(a)}
        \end{minipage}
        \begin{minipage}{0.4\textwidth}
            \centering
            \resizebox{\textwidth}{!}{
            \begin{tabular}{cccc}
                \toprule
                & \textit{Solo} & \textit{Handover} & \textit{Joint} \\
                \midrule
                Robot & 26.27 & \textbf{50.26} & 8.31 \\
                \midrule
                Pos. B & \textbf{37.95} & 17.51 & 15.70\\
                \midrule
                Pos. C & 2.39 & 3.64 & \textbf{50.3} \\
                \midrule
                Other & 45.04 & 48.31 & 22.90\\
                \bottomrule
            \end{tabular}}
            \label{tab:gaze_b}
            \textbf{(b)}
        \end{minipage}
        \\
        \vspace{10pt}
        \begin{minipage}{0.4\textwidth}
            \centering
            \resizebox{\textwidth}{!}{
            \begin{tabular}{cccc}
                \toprule
                & \textit{Solo} & \textit{Handover} & \textit{Joint} \\
                \midrule
                Robot & 9.68 & 15.48 & 3.69 \\
                \midrule
                Pos. B & 13.41 & 4.32 & 4.85\\
                \midrule
                Pos. C & 1.06 & 1.36 & 11.74 \\
                \midrule
                Other & 69.97 & 71.33 & 75.32\\
                \bottomrule
            \end{tabular}}
            \label{tab:gaze_all_before}
            \textbf{(c)}
        \end{minipage}
        \begin{minipage}{0.4\textwidth}
            \centering
            \resizebox{\textwidth}{!}{
            \begin{tabular}{cccc}
                \toprule
                & \textit{Solo} & \textit{Handover} & \textit{Joint} \\
                \midrule
                Robot & 17.45 & \textbf{33.61} & 16.65 \\
                \midrule
                Pos. B & \textbf{24.22} & 15.35 & 13.31\\
                \midrule
                Pos. C & 2.86 & 3.64 & \textbf{19.1} \\
                \midrule
                Other & 53.55 & 40.16 & 48.41\\
                \bottomrule
            \end{tabular}}
            \label{tab:gaze_all}
            \textbf{(d)}
        \end{minipage}
        \vspace{-7pt}
\end{table} 

For the handover action, the gaze percentage is higher at the robot. For non-handover actions, the gaze percentage is higher at the positions where participants would place the object (B for solo actions and C for joint actions). Before movement onset, as seen in Table~\ref{tab:gaze_a}(a), although the percentages suggest different behaviors for the three actions, these differences are not statistically significant. However, this trend becomes more pronounced after movement onset, as shown in Table~\ref{tab:gaze_b}(b) via the values in bold. Table \ref{tab:gaze_all_before}(c) and \ref{tab:gaze_all}(d) show the median percentages for all participants which also follow the aforementioned trend differentiating the three human actions. The existence of this trend before movement onset suggests that there is indeed some information in the gaze data that can be used to predict the type of upcoming human action before the motion begins. Since this trend becomes more pronounced after movement onset, it hints that the amount of information available in the gaze data increases as the action unfolds. This suggests that while initial predictions can be made using gaze data prior to the human motion onset, more accurate predictions can be made by utilizing gaze data as the human motion begins and progresses.
The percentages for individual participants are provided in the supplementary materials (Supplementary Tables 2 and 3).

\subsection*{EEG Data Analysis}
The motor cortex of the brain is in charge of movement-related functions such as movement planning and execution, which corresponds to the centrally located EEG channels in the 10-20 system \cite{yahya19}. Movement-related cortical potentials (MRCPs) and sensory-motor rhythms (SMRs) are key EEG characteristics during voluntary movement preparation and planning observed over the motor cortex \cite{sumeyra22}. 

\begin{figure*}[t]
    \centering
        \subfigure[]
   { \includegraphics[width=0.8\linewidth]{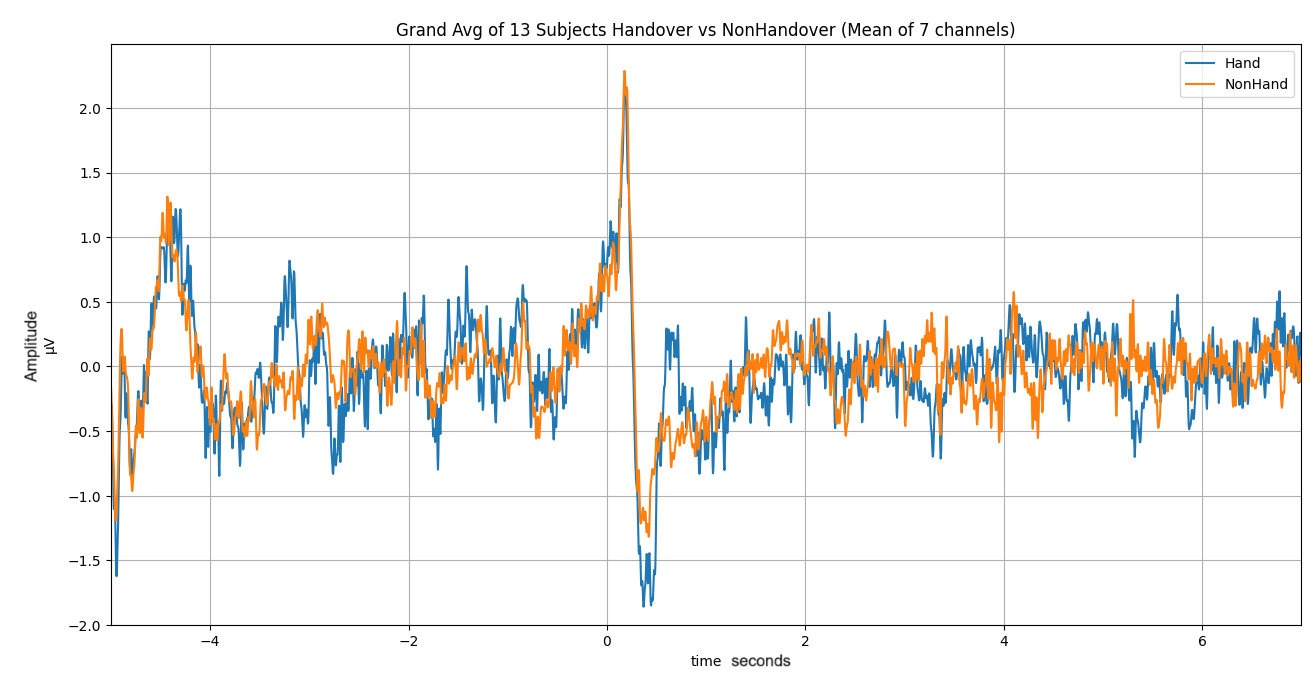} \label{fig:ERP} }
    \subfigure[]{\includegraphics[width=0.8\linewidth]{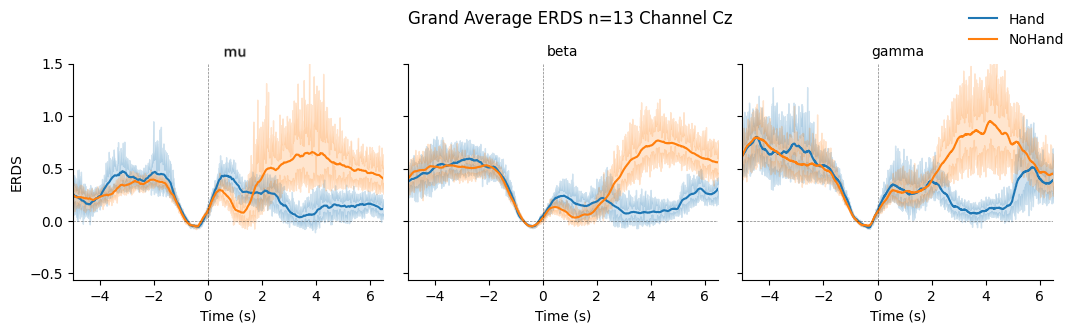}\label{fig:psd}}
        \caption{Grand average over 13 subjects for the two conditions, Blue: Handover, Orange: Non-Handover \textbf{(a)} ERP gathered from the motor cortex (Channels: C3, C4, Cz, CP1, CP2, FC1, FC2) \textbf{(b)} ERDS of mu, beta and gamma Channel Cz, with the variance highlighted by the shaded region.}
   \label{fig:erp_psd}
 \end{figure*}

MRCPs are event-related potentials (ERPs) that begin with a slow negative shift about two seconds before movement onset, turning positive with the movement \cite{olsen21}. The grand average ERPs for 13 subjects, comparing handover vs. non-handover conditions, are shown in Fig.~\ref{fig:ERP} for central channels C3, C4, Cz, CP1, CP2, FC1, and FC2. A negative trend starts around -1.8 seconds for both conditions, peaking negatively at -1.7 seconds, followed by a second negative peak at -0.7 seconds (Negative Slope) and finally a positive trend begins just before movement execution, peaking at 0.1 seconds (motor potential) which are in-line with what is observed in the literature \cite{olsen21}. MRCP peak and latency parameters are reported to differ with various movement types, task speed, whether the task is self-paced or cued, the uncertainty level, and the existence of force \cite{brunia2003cnv, nascimento2006movement, rektor2003intracerebral}.

SMRs are described as oscillations in the mu (8-12 Hz), beta (13-30 Hz), and gamma (31-200 Hz) frequency bands, seen either as a decrease in power called event-related desynchronization (ERD) or an increase in power called event-related synchronization (ERS). The mu, beta, and gamma band powers over the motor cortex decrease approximately two seconds before the onset with the movement preparation corresponding to ERD. The grand average of the relevant frequency band powers for channel Cz is shown in Fig.~\ref{fig:psd} calculated from $n=13$ subjects. It must be noted that since the EEG signals are band-pass filtered between 1-40 Hz during pre-processing, the gamma band contains frequencies between 30 and 40 Hz. ERD is observed between -2 and 0 seconds for both conditions in the mu, beta, and gamma bands as expected. A paired t-test showed that the band powers are significantly different between the handover and non-handover trials for most of the subjects (Supplementary Table 4).
The mu band is significantly different between handover and non-handover for all subjects except for S2 and S9, with a p-value $<0.05$. The beta band is significantly different for all 13 subjects in this time interval, meanwhile, the gamma band is significantly different for all subjects but one, S11, between the two conditions.

\section{Handover Intention Detection}
In this section, we compare the performance of gaze and EEG for handover detection against motion trajectories recorded by an RGBD camera, evaluating both detection timing and accuracy. We also assess the performance of single modalities versus multiple modalities. 

\subsection*{Model Selection}

The main goal of this study is to discriminate handover intention from other types of movements. Therefore, we combined data from the \emph{Solo} and \emph{Joint} conditions as class 0 and used data from \emph{Handover} trials as class 1. 

We compared the performance of two example classifiers: (1) linear classification using linear discriminant analysis (LDA), and (2) non-linear classification using a long-short-term memory (LSTM) neural network. LDA works by modeling the data distribution as a multivariate Gaussian distribution. It then uses the Bayes theorem to classify new data points. 

We used the Scikit-Learn package~\cite{scikit-learn} in Python to implement the linear classification pipeline. LDA was applied using Scikit-Learn’s default parameters. To evaluate LDA’s performance, we used repeated stratified 10-fold cross-validation with three repeats. This involved splitting the data into 10 folds, using 9 folds for training and 1 fold for testing, and repeating the procedure so each fold served as the test set. This entire process was repeated three times with different randomizations.

For the neural network classifier, we used the PyTorch package~\cite{paszke2019pytorch} in Python to implement data loading, training, and the LSTM model. Different LSTM parameters were used for EEG data compared to gaze and hand motion trajectory data due to EEG's higher dimensionality. Specifically, for EEG, we used a 1-layer LSTM with 128 hidden units. The LSTM's output for the last time step was passed through a ReLU activation function and then mapped to class probabilities using a fully connected layer followed by a Sigmoid function. We used a batch size of 16 and trained the model for 100 epochs, selecting the model with the minimum validation loss for evaluation on the test split. Given the low number of data samples, we aimed to keep the hidden layers of LSTMs to a minimal number and correspondingly selected batch size and other parameters. For the gaze and the motion data, we used a 2-layer LSTM with 10 hidden units while using a batch size of 5 and training for 200 epochs with early stopping based on validation loss after 100 epochs. Therefore, we selected the model with the minimum validation loss for further evaluation on the test split.

To select the best neural network model based on validation loss, we included both a validation set for model selection and a test set for final evaluation. We employed a nested cross-validation approach to achieve this. Specifically, we used repeated stratified 10-fold cross-validation with three repeats to choose test splits. For each of the nine training folds, we performed an additional 10-fold cross-validation to train the model on various train-validation splits. Finally, a weighted ensemble of these 10 models was used to evaluate the LSTM's performance on the test split.

\subsection*{Performance Evaluation}

We measured the area under the receiver operating characteristic curve (AUC-ROC) to evaluate the classification performance for the classifier models. The ROC curve is obtained by using different threshold values on the model's predictions and computing the true positive rate (TPR) and false positive rate (FPR) given by equations~\ref{eq:tpr} and~\ref{eq:fpr} respectively. 

\begin{equation}
    \label{eq:tpr}
    TPR = \frac{TP}{TP + FN} ,
\end{equation}

\begin{equation}
    \label{eq:fpr}
    FPR = \frac{FP}{FP + TN} ,
\end{equation}

TP, TN, FP, and FN stand for true positive, true negative, false positive, and false negative samples. Plotting TPR against FPR creates the ROC curve. AUC-ROC is by definition the area under the ROC curve. An AUC-ROC greater than 0.5 indicates that the classifier is performing better than a random classifier. An AUC-ROC of 1.0 shows a perfect classification discriminating all positive samples from the negative ones. 

\subsection*{Feature Extraction}

\subsubsection*{EEG Data}

We applied time-frequency transformation to the EEG data using Morlet wavelets with three cycles to generate input for both classifiers. This transformation covered a frequency range of 5–40 Hz within the selected time window, encompassing the alpha (8–12 Hz), beta (13–30 Hz), and part of the gamma (\(>\text{30 Hz}\)) bands, which are known to encode motor task information in EEG signals~\cite{amo2020induced, yu2022effects}. Frequencies above 40 Hz were excluded due to susceptibility to high-frequency line noise and muscle artifacts. The time-frequency representations were then averaged over channels and reshaped into 1-D vectors suitable for the linear classifier. Before applying LDA, we standardized the data to zero mean and unit standard deviation and used principal component analysis to extract components that capture 99\% of the data variance. Applying PCA before LDA further reduces the dimensionality of the data and helps to regularize the classification problem. We also standardized the EEG data for LSTM training, although no further dimensionality reduction (e.g. by PCA) was applied.  
 
\subsubsection*{Gaze Data}
The features for the gaze data were the X and Y coordinates of the recorded gaze data for the participants, referenced to a fixed point on the robot's torso (Supplementary Fig. 3). These features are denoted by: $GazeX$ and $GazeY$. 

\subsubsection*{Hand Motion Trajectory Data}

The features for the motion data were the X, Y, and Z coordinates of the tracked right hand, denoted by: $HandX$, $HandY$, $HandZ$. These coordinates are referenced to the Azure Kinect camera, which was mounted on the torso of the robot (Fig.~\ref{fig:experimental setup}).



\begin{figure}[h!]
    \centering
    \subfigure[]{\includegraphics[width=0.3\linewidth]{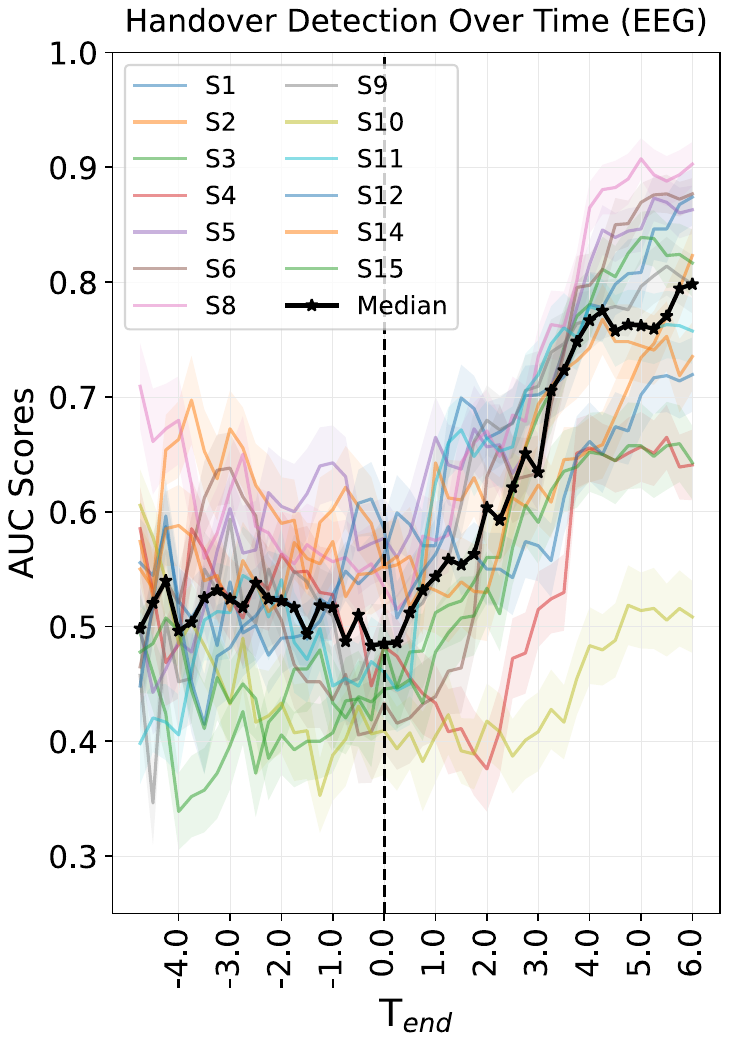}\label{fig:eeg_lda}}
    \subfigure[]{\includegraphics[width=0.3\linewidth]{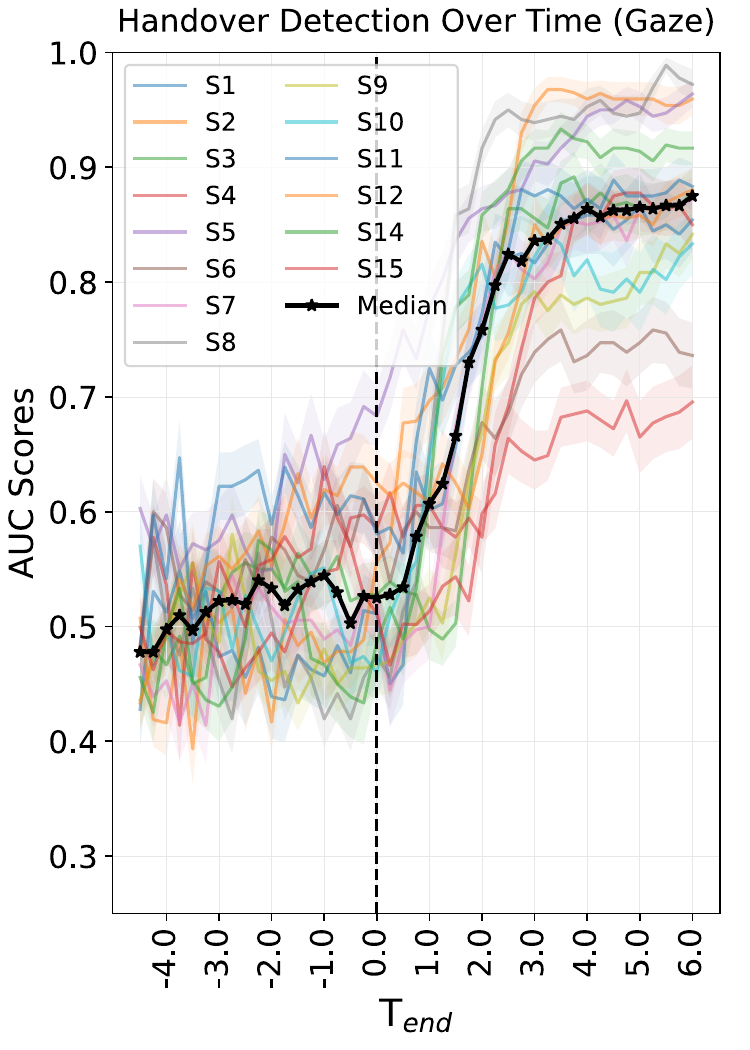}\label{fig:gaze_lda}}
    \subfigure[]{\includegraphics[width=0.3\linewidth]{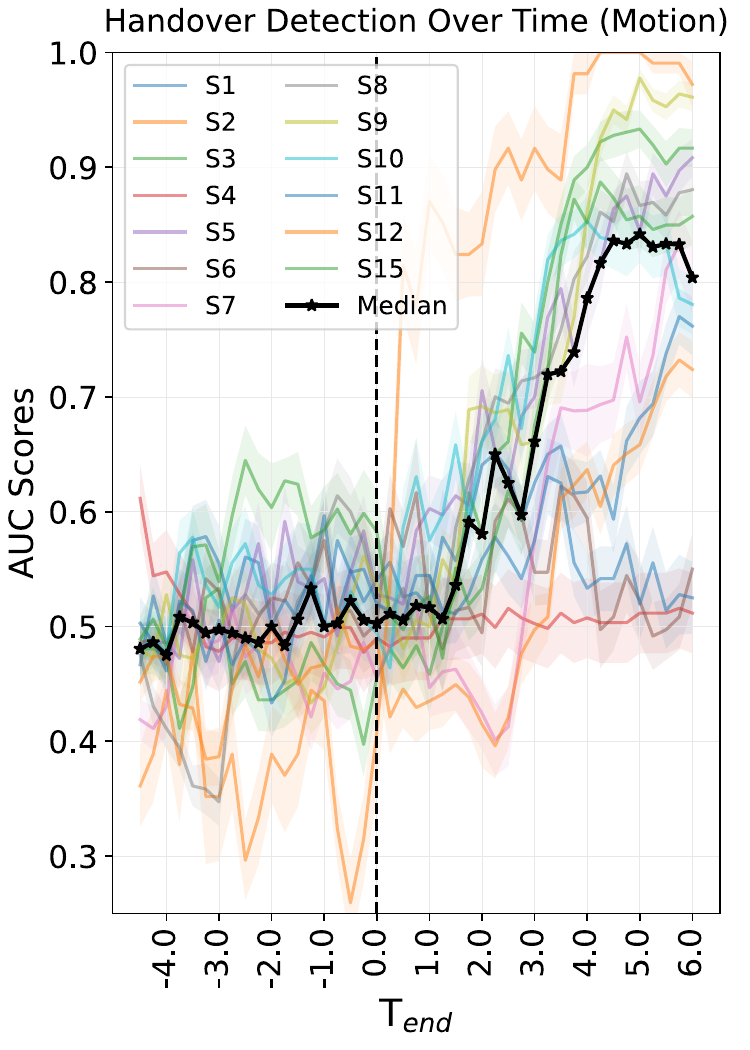}\label{fig:motion_lda}}
    \subfigure[]{\includegraphics[width=0.3\linewidth]{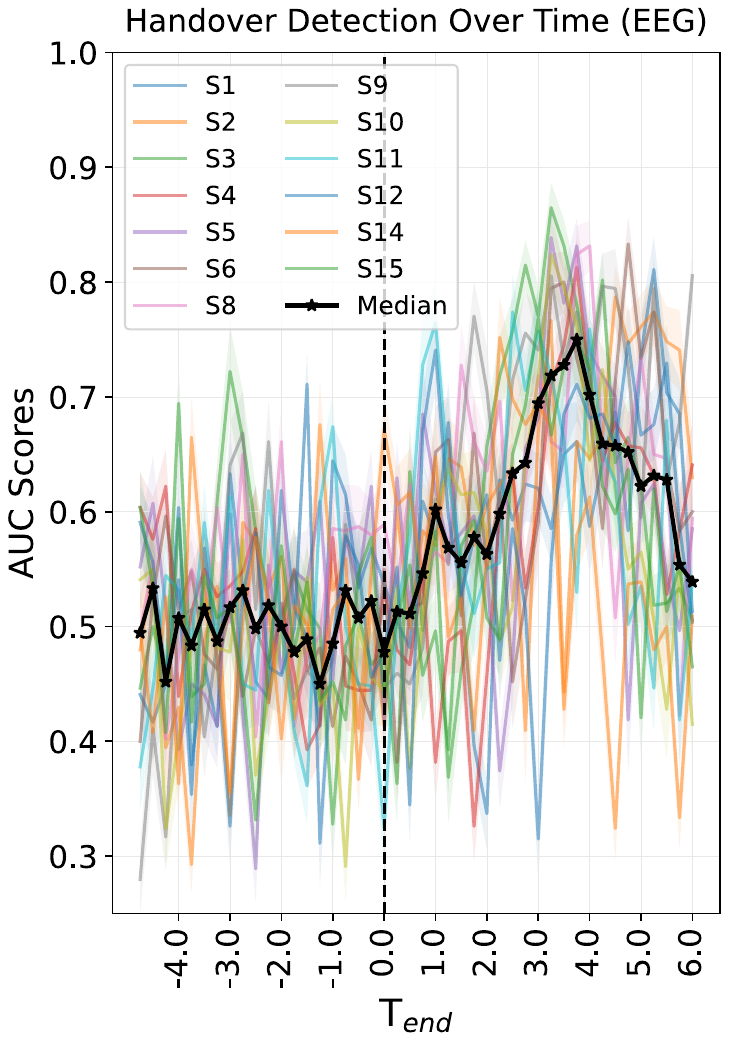}\label{fig:eeg_lstm}}
    \subfigure[]{\includegraphics[width=0.3\linewidth]{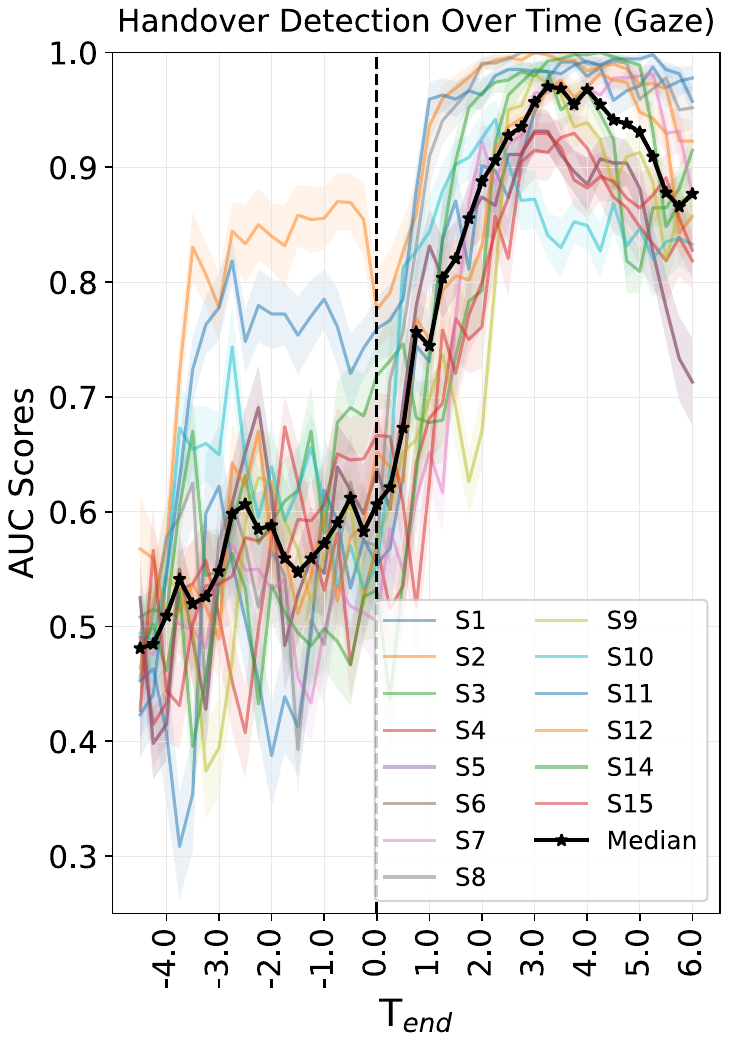}\label{fig:gaze_lstm}}
    \subfigure[]{\includegraphics[width=0.3\linewidth]{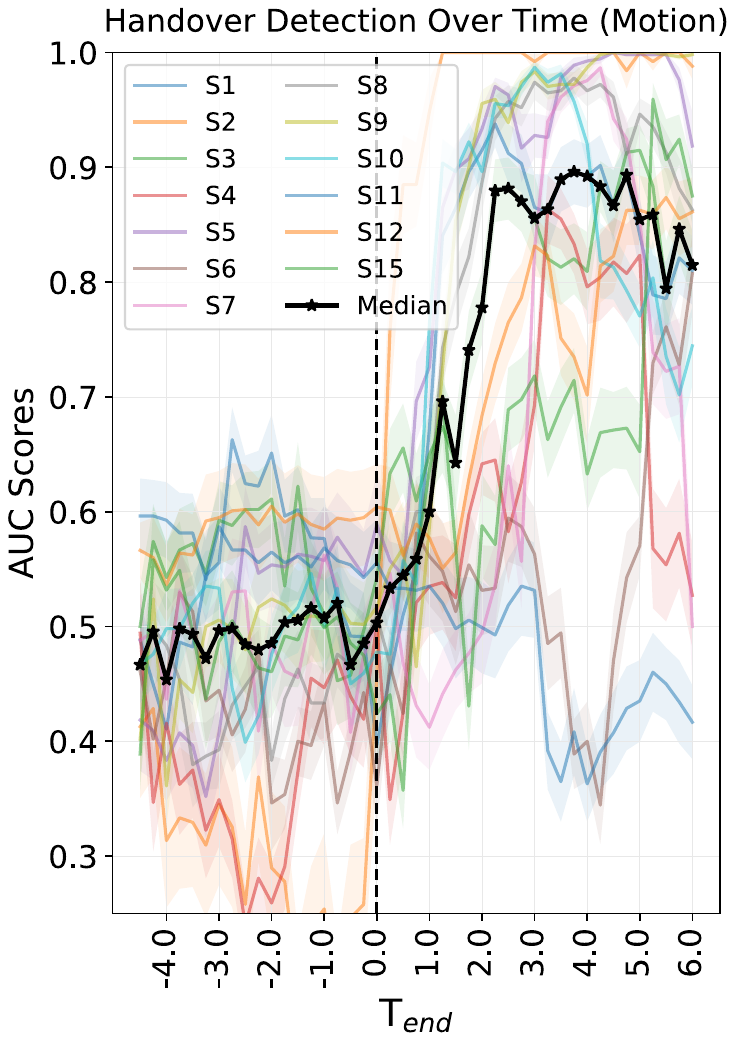}\label{fig:motion_lstm}}
    \caption{\justifying Detecting Handover Intention, Results of LDA (a, b, c) and LSTM (d, e, f) training on increasing time windows for (a, d) EEG, (b, e) gaze, and (c, f) hand motion. Different colors show individual participants, with the black line as the median performance and shaded areas as the standard error. The x-axis shows the end time of the training window and the dashed line at 0.0 marks movement onset.}
    \label{fig:subj-lda-lstm}
\end{figure}

\subsection*{Detecting Handover Intention Based on Single Modalities} 

To compare the effectiveness of EEG, gaze, and hand motion trajectory modalities in detecting human-to-robot handovers, we trained a linear classifier using linear discriminant analysis (LDA) for each participant on data from individual modalities across increasing time windows. Starting from the cue onset (-5 seconds), we used time windows incremented by 250 ms up to 6 seconds after movement onset (-5 to -4.75, -5 to -4.5, ..., -5 to 6 seconds). The linear classifier was trained and tested on each time window separately, with the results presented in Fig.~\ref{fig:eeg_lda}~\ref{fig:gaze_lda} and~\ref{fig:motion_lda}. 
Our findings show that detection performance improves significantly for all modalities after the movement begins. However, gaze data shows an earlier increase in performance compared to EEG and hand motion trajectory. Notably, gaze data even demonstrates some predictive power before movement onset, indicating participants' movement planning phase. Although EEG performance starts its rise from 50\% earlier than the hand motion trajectory (0.75 s vs. 1.25 s after movement onset), The hand motion performance reaches higher values faster than EEG. 


\begin{figure}[b!]
    \centering
    \subfigure[]{\includegraphics[width=0.49\textwidth]{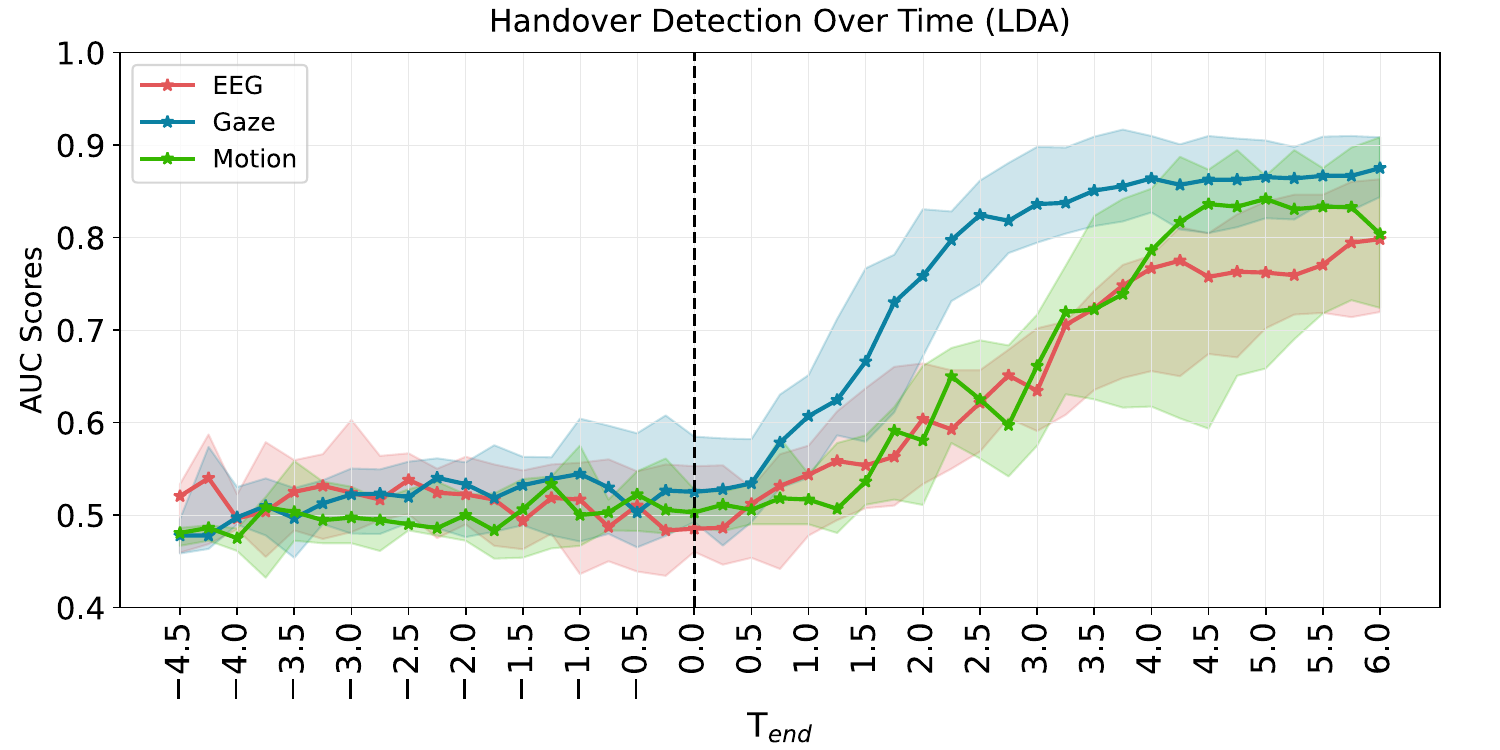}\label{fig:all_lda}}
    \subfigure[]{\includegraphics[width=0.49\textwidth]{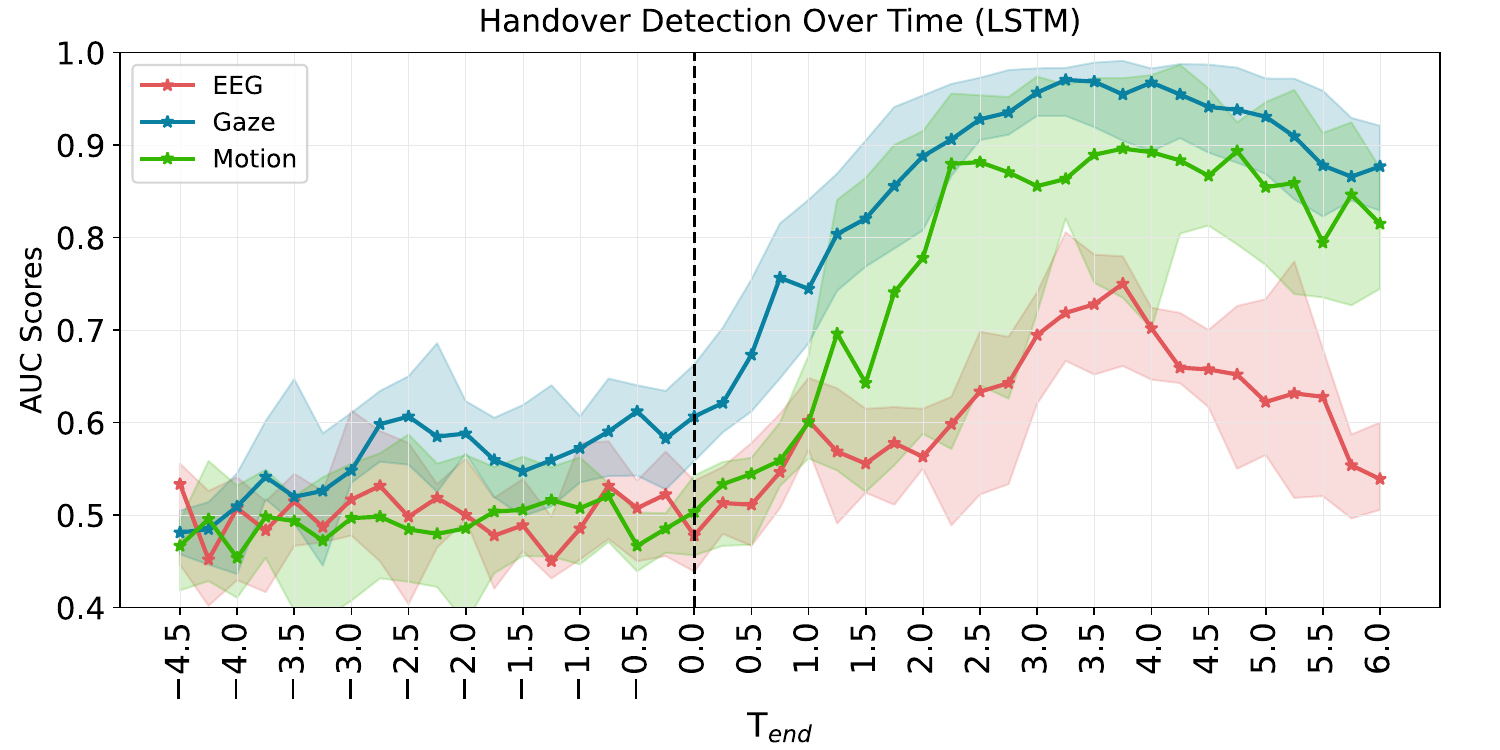}\label{fig:all_lstm}}
    \caption{\justifying Comparing the handover detection performance of (a) LDA and (b) LSTM models across modalities. The graphs show the median performance for all modalities, with the shaded area indicating the 25th and 75th percentiles.}
    \label{fig:lda-lstm}
\end{figure}
\begin{table}[h!]
    \caption{Time (in seconds) at which different AUC-ROC levels are reached for the median performance of each modality. An AUC-ROC level is considered reached if it is sustained for at least three consecutive time steps; ``X" indicates the AUC-ROC level was not reached. Values in bold mean a significant difference was seen in values for one-way Anova analysis}
    
        \resizebox{0.45\textwidth}{!}{
        \begin{tabular}{cccc}
            \toprule
            \multirow{2}{*}{AUC Levels} & \multicolumn{3}{c}{Time in secs} \\
            \cmidrule{2-4} LDA
             & EEG & Gaze & Hand-Motion \\
            \midrule
            0.60 & 3.0 & 1.5 & 3.5\\
            0.65 & 3.75 & 2.0 & 3.5\\
            0.70 & 3.75 & \textbf{2.25} & 3.75\\
            0.75 & 4.5 & \textbf{2.5} & 4.5 \\
            0.80 & X & \textbf{3.0} & 4.75\\
            0.85 & X & \textbf{4.0} & X \\
            0.90 & X & X & X \\
            \bottomrule
        \end{tabular}
        }
        \label{tab:performance_lda}
    \hspace{0.001\textwidth}
        \resizebox{0.45\textwidth}{!}{
        \begin{tabular}{cccc}
            \toprule
            \multirow{2}{*}{AUC Levels} & \multicolumn{3}{c}{Time in secs} \\
            \cmidrule{2-4} LSTM
             & EEG & Gaze & Hand-Motion \\
            \midrule
            0.60 & 3.5 & \textbf{0.5} & 1.5\\
            0.65 & 3.0 & \textbf{1.0} & 2.25\\
            0.70 & X & \textbf{1.25} & 2.25\\
            0.75 & X & \textbf{1.75} & 2.5 \\
            0.80 & X & 1.75 & 2.75\\
            0.85 & X & 2.25 & 2.75\\
            0.90 & X & 2.75 & X \\
            \bottomrule
        \end{tabular}
        }
        \label{tab:performance_lstm_lda}
    
\end{table}
To quantify the differences between the median performances of each modality across participants, we identified the earliest time each modality reached a specific AUC-ROC level and maintained it for at least three consecutive time points. The results are summarized in Table~\ref{tab:performance_lda}. Our findings indicate that the gaze modality achieved the highest AUC-ROC among the three. Additionally, gaze data predicted the intention of handover faster than the other modalities, sometimes up to 1.75 to 2 seconds earlier than EEG or hand tracking (Table~\ref{tab:performance_lda}). Interestingly, EEG was neither faster nor more accurate than hand motion, as it did not achieve the same high AUC-ROC levels (Table~\ref{tab:performance_lda}).

We repeated the experiment using a neural network classifier, specifically a long-short-term memory (LSTM) network~\cite{hochreiter1997long}, which is well-suited for sequence data like time series classification. The results, shown in Fig.~\ref{fig:eeg_lstm}~\ref{fig:gaze_lstm} and~\ref{fig:motion_lstm} and Table \ref{tab:performance_lstm_lda}, indicate that the LSTM outperforms the LDA classifier for both gaze and hand motion trajectory modalities. However, for EEG signals, the LSTM's performance is highly variable over time and across subjects, suggesting it failed to learn robust patterns from the data. We hypothesize this is due to the high-dimensional nature of EEG data, which demands more samples for reliable training. In this situation, neural networks are more prone to overfitting the noise in the data due to their large number of parameters. Fig.~\ref{fig:lda-lstm} displays the median participant performance of LDA and LSTM trained on different modalities in one plot.

\subsection*{Multimodal Handover Detection}
\begin{figure*}[t!]
    \centering
    \subfigure[]{\includegraphics[width=0.4\linewidth]{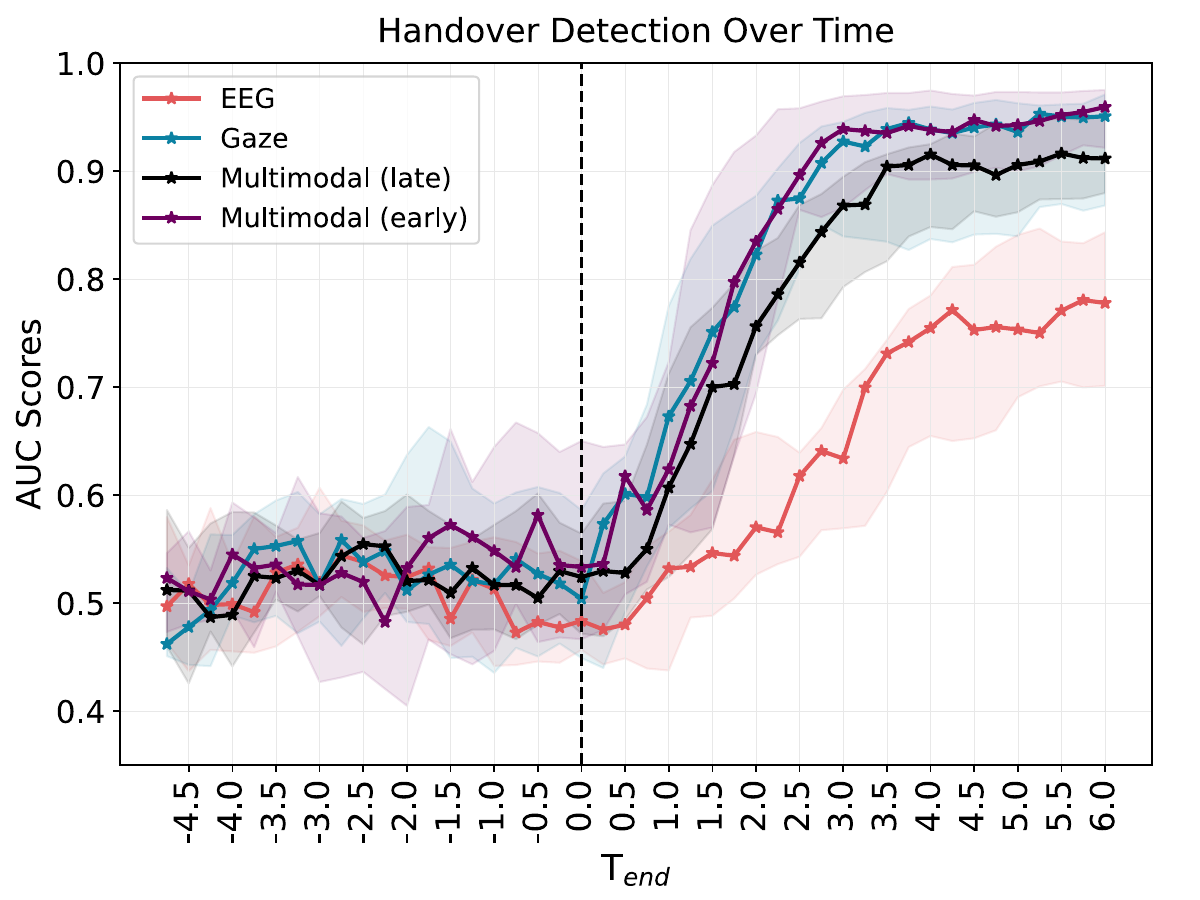}\label{fig:multimodal_eeg-gaze}}
    \subfigure[]{\includegraphics[width=0.4\linewidth]{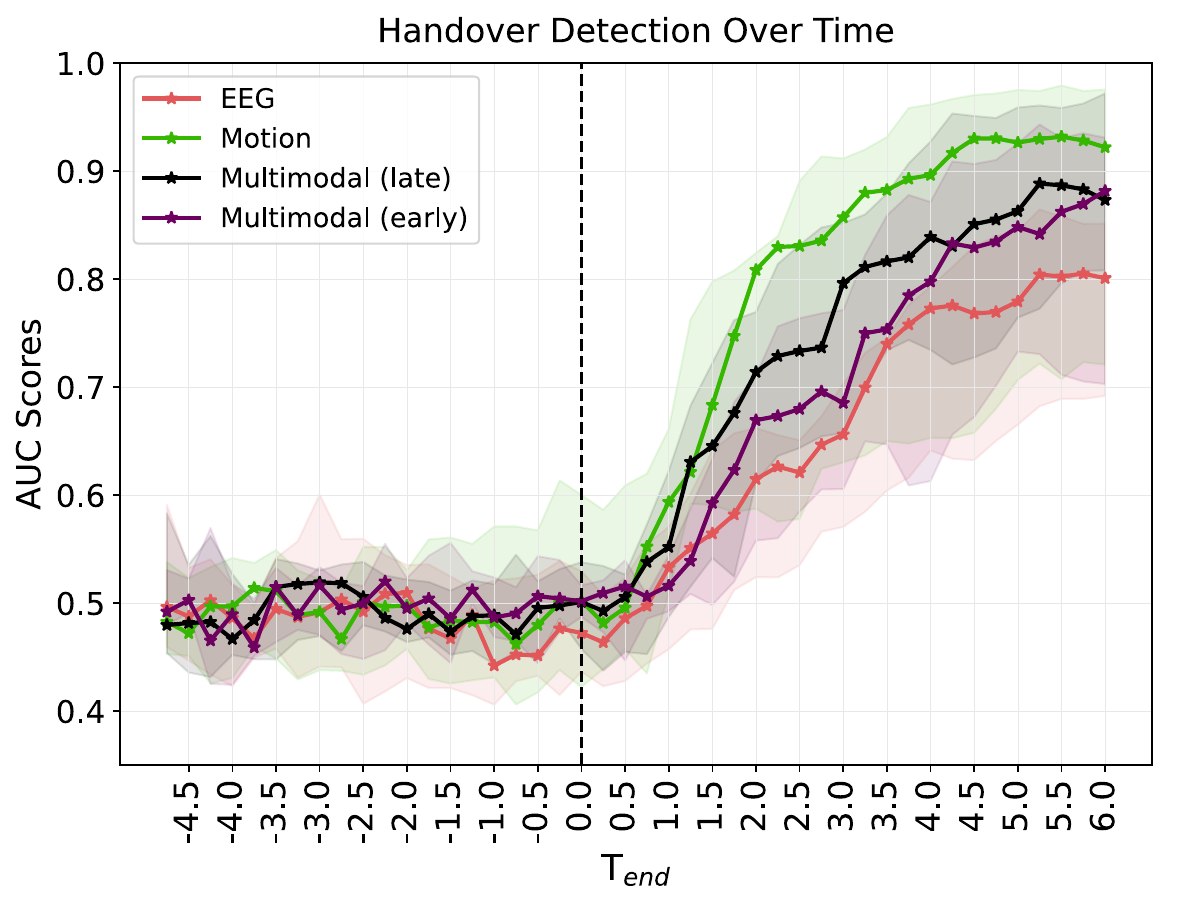}\label{fig:multimodal_eeg-motion}}
    \subfigure[]{\includegraphics[width=0.4\linewidth]{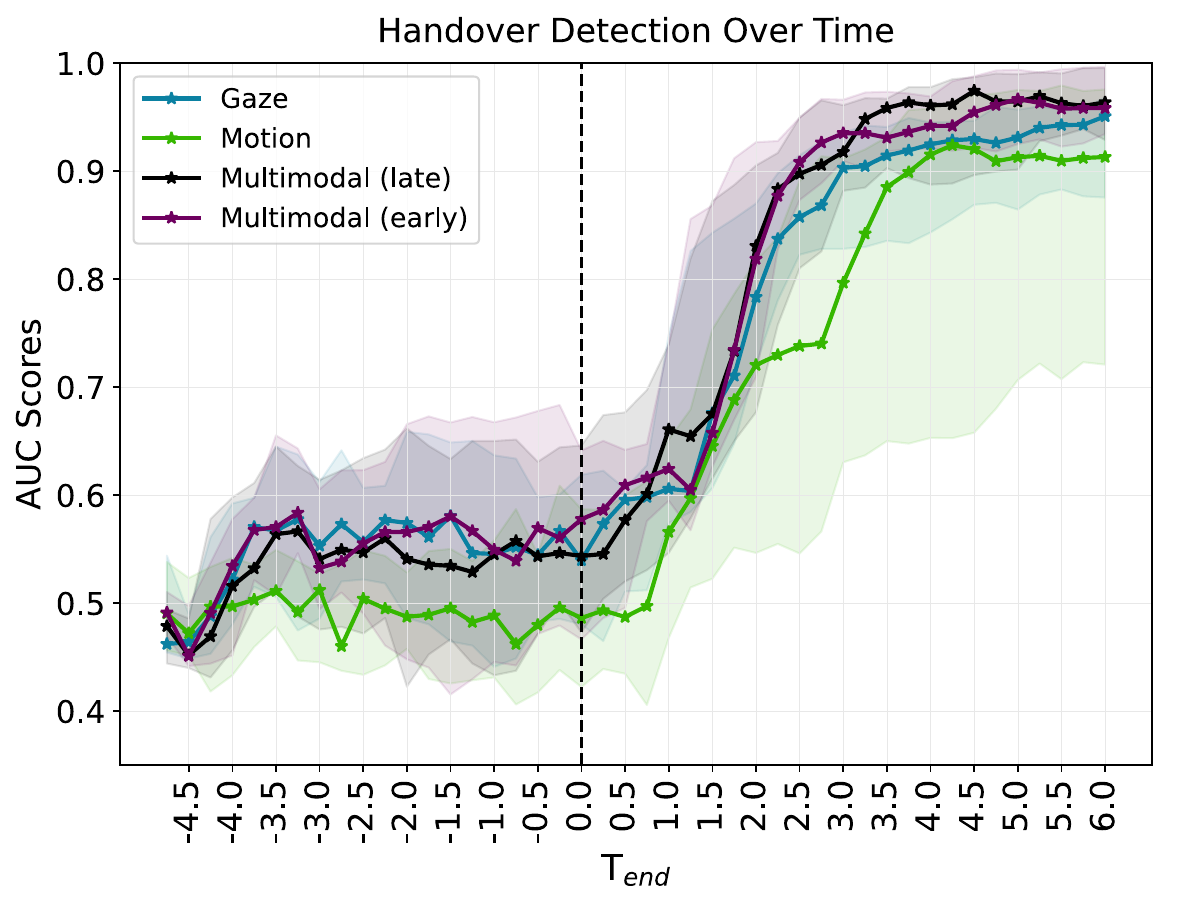}\label{fig:multimodal_gaze-motion}}
    \subfigure[]{\includegraphics[width=0.4\linewidth]{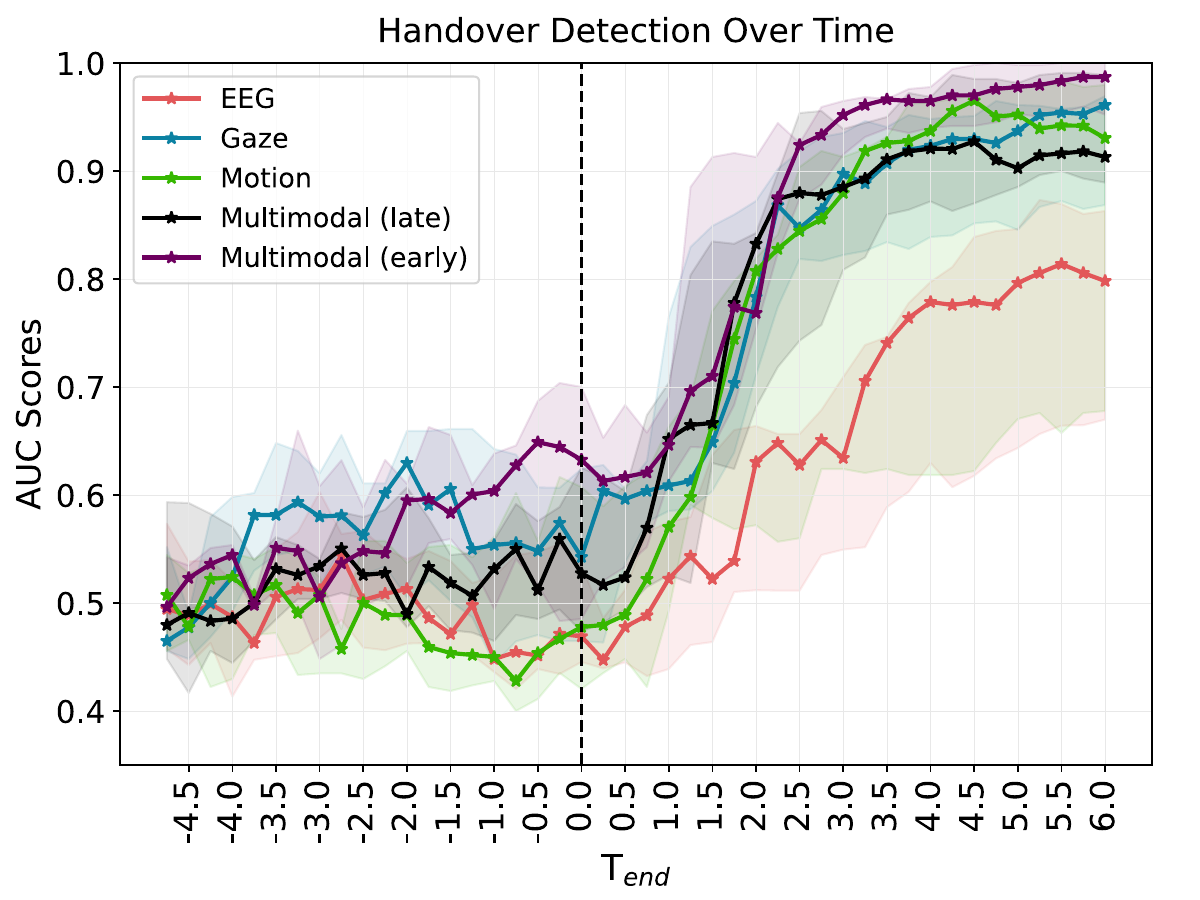}\label{fig:multimodal_all}}
    \caption{A comparison between LDA classifiers trained on single modalities versus multiple modalities using early or late fusion. (a) is based on EEG and gaze, (b) is EEG and hand motion, (c) is a gaze and hand motion, and (d) involves all three modalities. Each graph shows the median performance over subjects with the shaded area indicating the interquartile range. Note that a different set of participants is considered in each subplot based on the number of available trials per individual.}
    \label{fig:multimodal}
\end{figure*}

After assessing the predictive power of individual modalities for handover detection, we examined whether integrating these modalities into a multimodal classifier could enhance the performance. Using LDA as the classification method, we considered both (1) early fusion and (2) late fusion of different modalities. For early fusion, we concatenated data from all involved modalities into a single vector for the LDA model. For late fusion, we averaged the class probabilities from classifiers trained on each modality, weighted by their training performance, to generate new class probabilities. We then computed the AUC-ROC for these combined probabilities. The results of both approaches are shown in Fig.~\ref{fig:multimodal}. The results indicate that gaze is the most influential modality, enhancing the performance of both EEG and hand motion trajectory when used in a bimodal model (Fig.~\ref{fig:multimodal_eeg-gaze}, \ref{fig:multimodal_gaze-motion}). However, this enhancement is constrained by the inherent performance of the gaze data alone. Combining all three modalities with the early fusion approach significantly improves performance over each individual modality (Fig.~\ref{fig:multimodal_all}). Notably, this hybrid model can predict handover intention better than chance around one second before the expected movement onset at 0, although the performance is still not very high. 

The multimodal model was only trained for participants who had at least 60 uncorrupted trials of the modalities involved in training (Supplementary Table S3). Therefore, each of the four subplots in Fig.~\ref{fig:multimodal} may be based on a different set of subjects. Also for early fusion involving EEG, we first reduced the dimensionality of EEG signals using principal component analysis (PCA) and then concatenated the result with other modalities. This was to make the dimensionality of EEG signals comparable with that of other modalities.

\begin{figure}[b!]
    \centering
    \subfigure[]{\includegraphics[width=0.44\linewidth, height=0.5\linewidth]{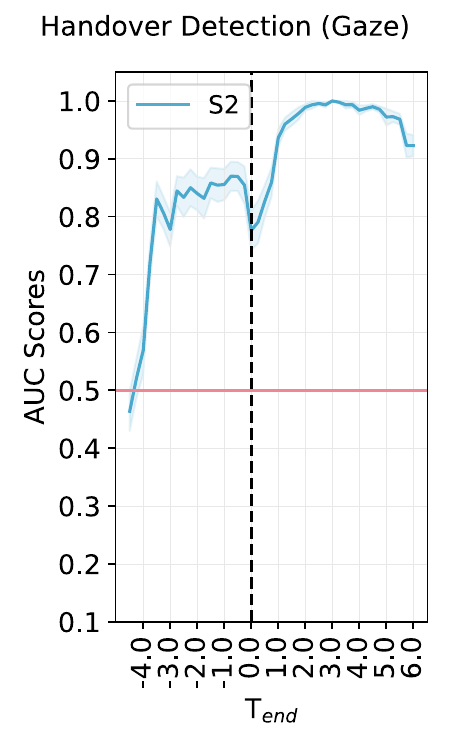}\label{fig:selected_gaze}}
    \subfigure[]{\includegraphics[width=0.44\linewidth, height=0.5\linewidth]{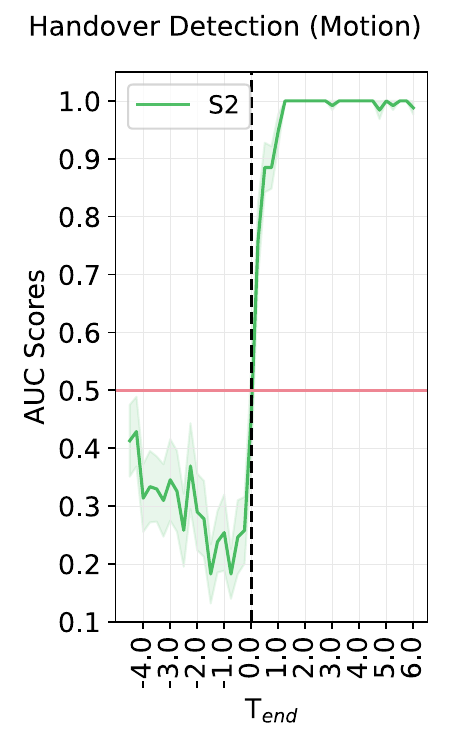}\label{fig:selected_motion}}
    \subfigure[]{\includegraphics[width=0.9\linewidth, height=0.52\linewidth]{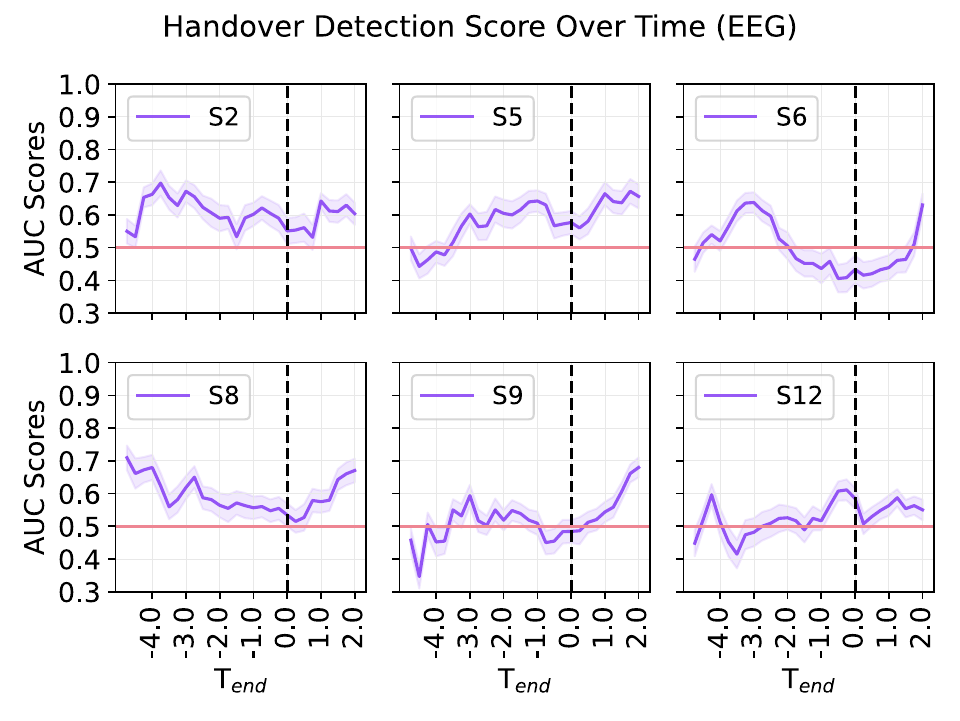}\label{fig:selected_eeg}}
    \caption{AUC-ROC score over time for handover intent classification for selected participants from different modalities: (a) LSTM trained on gaze data from participant 2, (b) LSTM trained on hand motion data from participant 2, (c) LDA trained on EEG data from participants 2, 5, 6, 8, 9, and 12.}
    \label{fig:selected}
\end{figure}




\section{Discussion}

In this work, we examined how different modalities predict human motion intention in close-proximity human-robot collaborations, focusing on distinguishing human-to-robot handovers from other tasks. We compared human gaze and EEG data with hand motion trajectories, our baseline modality for detecting handovers. Our results show that hand motion trajectories can identify a handover about 1 second after motion onset. Notably, the human gaze can detect handover intention earlier than both EEG and hand motion trajectories, sometimes even before the movement begins, suggesting it could enhance conventional detection methods. In contrast, EEG performance had more variation across subjects and trials. Although EEG could predict object handover from the other two conditions, it was less accurate compared to the other two modalities. 

Comparing AUC-ROC levels in LSTM and LDA models, gaze data achieved the highest AUC-ROC levels after movement onset (Table~\ref{tab:performance_lstm_lda}), and it reached these levels faster than both hand motion and EEG, demonstrating its superior predictive power. For some participants
(Fig.~\ref{fig:selected_gaze}), a high AUC-ROC was observed even before movement onset, highlighting gaze data's predictive power prior to actual movement. These results align with earlier analysis describing percentages of gaze data before movement onset (\ref{tab:gaze_b}).

In terms of motion data, LDA classification performance post-onset was similar to that of EEG. However, median plots in Fig.~\ref{fig:lda-lstm} show that motion data's predictive power for handover intention is lower than that of gaze data, with no predictive power before motion onset. For participant 2 (Fig.~\ref{fig:selected_motion}), there was a steep rise in AUC-ROC just after motion onset, surpassing gaze in predictive power, a result not seen in other participants.

Our initial analysis of averaged EEG trials indicated the feasibility of detecting handover intention before movement onset. However, single-trial EEG signals could not robustly predict handover intention before movement began. While EEG classification results showed some predictive power for certain participants, this varied across individuals. Fig.~\ref{fig:selected_eeg} shows LDA EEG classification results for selected participants before movement onset, highlighting some predictive power (up to 70\% AUC-ROC) for handover detection, albeit at varying times. The low signal-to-noise ratio of EEG signals likely contributed to poor performance. Our classification results suggest a need for more EEG trials to help models identify relevant patterns despite signal artifacts.

We also explored whether combining different modalities could improve detection performance. Specifically, we investigated if different modalities provide complementary information about the handover task, enhancing the model's certainty regarding the type of upcoming motion. Our findings indicate that gaze data mainly drives the model's performance. However, the early fusion approach using all three modalities (Fig.~\ref{fig:multimodal_all}) outperformed any single modality right before movement onset. The EEG-gaze model (Fig.~\ref{fig:multimodal_eeg-gaze}) also showed a performance boost right before movement onset, though less pronounced, suggesting that EEG might add valuable information when combined with gaze data. Conversely, combining EEG with motion data worsened performance compared to using motion data alone, indicating that EEG may introduce more uncertainty into the decision-making process.

Some limitations in this study should be considered when interpreting the results. First, we only considered a controlled and limited set of human-robot collaboration scenarios, with identical initial actions across conditions, differing only by human intent. This design aimed to create a challenging scenario for detecting human intention, which could lead to robot failure. However, real-world applications often involve diverse task variations that might provide additional cues for the robot to understand human intention earlier and more accurately. Furthermore, our experiment included only reactive tasks, where human actions responded to an external cue, differing from real-world scenarios where humans decide the motion type and timing, potentially altering brain responses.

Another limitation is measuring the exact time participants started the movement due to the reactive nature of the design. Although we used the green \emph{Go!} signal as the movement onset cue, participants often began moving slightly later due to natural reaction times, sometimes up to one second after the signal. Thus, the \emph{Go!} signal marks the earliest possible start time rather than the actual movement start. This does not affect our claims, as our main goal is to compare the timing of different modalities relative to each other.

The number of trials per participant is also a limiting factor in many experiments involving human subjects, as human fatigue requires keeping experimental sessions short. This is particularly problematic in repetitive tasks and when using devices like EEG that can become uncomfortable over time. Nevertheless, both the number of participants and the number of trials per participant are in line with the literature~\cite{cooper20, BUERKLE2021}. 

In this study, we collected simultaneous EEG, gaze, and hand motion trajectory data. Among these modalities, hand motion trajectory is the most convenient for the user, requiring only an RGBD camera installed on the robot and posing no extra burden on the user. Gaze data was collected using specialized wearable glasses for a more accurate tracking, although it can also be tracked using an RGB camera~\cite{Open_gaze}, similar to motion tracking. On the other hand, EEG data collection with currently available devices is the most challenging. We used wet EEG electrodes that require a conductive gel to be injected under the electrode on the scalp surface. While dry electrodes are available, they are more susceptible to noise. Additionally, recording EEG requires participants to minimize head movement to prevent muscle and electrode displacement artifacts, introducing inconvenience and reducing the likelihood of its widespread adoption in real-life scenarios. However, recent advances in recording hardware and signal processing methods may help mitigate these challenges.

Since EEG signals are recorded from the scalp's surface, they are highly susceptible to various noises and artifacts, such as line noise, muscle activity, blinking, and electrode displacement. Consequently, extensive preprocessing is often necessary to enhance the signal-to-noise ratio. However, certain preprocessing steps, like frequency filtering, can introduce temporal delays to the signal, which can affect the timing of movement intention detection.

It is to be noted that gaze and motion tracking too come with some challenges. Both these modalities require image processing in their preprocessing steps and thus suffer from the loss of tracking accuracy and tracking data due to several factors such as occlusion, lighting variations, and sensor noise.
In our analysis, we also suffered from the loss of samples in the data stream from gaze and motion.
The complexity of accurately interpreting visual data can lead to tracking errors, especially in dynamic or cluttered environments. Furthermore, the computational demands of real-time image processing may strain system resources, potentially resulting in delays or decreased performance.

The findings indicate that for robot controllers, the gaze data of humans holds the most potential for real-time prediction of human intention for handovers. 
Its high predictive power to determine if human motion is intended for handover just after the start of such motion is critical for creating a responsive system. It also shows some such predictive power just before the human motion starts and could be used to prepare the robotic system for a suitable response.
While hand motion tracking is convenient and widely used for recognizing human intention in human-robot handovers, its delayed prediction limits its usefulness in such dynamic interactions requiring precise timing. These delays could potentially cause a delayed response from the robot to the human motion, leading to an unnatural and awkward interaction experience. Despite its potential, as seen in the EEG data analysis, the EEG data is noisy and variable, making it unreliable for immediate practical application for such a prediction of human intention. 
For a strong controller, a combination of gaze data and refined hand motion analysis, supplemented by advanced signal processing techniques to address EEG data challenges, could improve intention prediction and enable timely, efficient robot responses in real-time scenarios.

Future work will involve integrating the assessed modalities into an online closed-loop system, where a robot controller predicts human intentions based on real-time sensory data and responds accordingly. Building such an online system requires real-time data processing, presenting new challenges. Additionally, the model's uncertainty must be considered when inferring human intentions.

For EEG signals, collecting more data can help models learn better representations. Given the high dimensionality of EEG data, we either need extensive feature extraction to reduce dimensionality and eliminate irrelevant components, or we require a large number of samples to allow more complex models, such as deep neural networks, to effectively handle the data.

Finally, we want to highlight the potential of foundation models and their high performance across various downstream tasks. These models have shown promising results in many human-robot interaction scenarios~\cite{min2022rethinking,mahadevan2024generative,lin2023gesture}. However, training and fine-tuning them usually require a large amount of data. Additionally, these models are so large that even inference demands significant memory resources. Current foundation models are well-developed for vision and language inputs, but incorporating other modalities such as gaze and EEG will require adapting the existing models to these modalities, necessitating further training and data collection. Future work would also evaluate the feasibility of using foundation models in the context of human-robot collaboration and specifically handovers.

\section*{Acknowledgment}
This work was supported by Digital Futures at KTH,
the Swedish Foundation for Strategic Research (BOS), Knut
and Alice Wallenberg Foundation, the Swedish Research
Council and ERC BIRD 884807.

\bibliographystyle{elsarticle-num-names} 
\bibliography{jrnl}

\section*{Supplementary Material}

\captionsetup[table]{labelfont=bf,labelformat=simple,labelsep=space,name=Supplementary Table}
\captionsetup[figure]{labelfont=bf,labelformat=simple,labelsep=space,name=Supplementary Figure}

\subsection*{Supplementary Tables}

\subsection*{}
\textbf{Included Trials Per Modality}
\begin{table}[h]
    \centering
    \caption{Gaze and Motion Data for Different Participants}
    \begin{tabular}{cccc}
        \toprule
        \textbf{Participant} & \textbf{Gaze} & \textbf{Motion} & \textbf{EEG}\\
        \midrule
        1  & 59 & 78 & 90 \\
        \midrule
        2  & 80 & 29 & 90 \\
        \midrule
        3  & 90 & 90 & 90 \\
        \midrule
        4  & 90 & 61 & 90\\
        \midrule
        5  & 90 & 90 & 90 \\
        \midrule
        6  & 90 & 90 & 90\\
        \midrule
        7  & 90 & 75 & -\\
        \midrule
        8  & 90 & 90 & 90\\
        \midrule
        9  & 90 & 90 & 90\\
        \midrule
        10 & 89 & 90 & 90\\
        \midrule
        11 & 90 & 90 & 90\\
        \midrule
        12 & 90 & 89 & 90\\
        \midrule
        13 & 90 & - & -\\
        \midrule
        14 & 90 & - & 90\\
        \midrule
        15 & 82 & 70 & 90 \\
        \hline
    \end{tabular}
    \label{table:gaze_motion_eeg data}
\end{table}

\subsection*{}
\textbf{Percentages of Gaze in Different Regions}
\begin{table}[H]
    \centering
    \caption{Percentages of Gaze, Time t=-5 to +3 s, before and after movement onset}
    \label{tab:participant8}
    \begin{minipage}{0.4\textwidth}
        \centering
        \small
        \resizebox{\textwidth}{!}{
        \begin{tabular}{|c|c|c|c|}
            \hline
            & Solo & Handover & Joint \\
            \hline
            Robot & 11.88 & 28.97 & 7.34\\
            \hline
            Pos. B & 30.11 & 28.87 & 17.64\\
            \hline
            Pos. C & 4.46 & 2.03 & 35.54\\
            \hline
            Other & 53.55 & 40.14 & 42.48\\
            \hline
        \end{tabular}}
        \captionof{subtable}{For participant 1}
        \label{tab:gaze_b1}
    \end{minipage}
    \hspace{35pt}
    \begin{minipage}{0.4\textwidth}
        \centering
        \small
        \resizebox{\textwidth}{!}{
        \begin{tabular}{|c|c|c|c|}
            \hline
            & Solo & Handover & Joint \\
            \hline
            Robot & 12.08 &  49.57 & 21.81 \\
            \hline
            Pos. B &  25.28 & 5.73 & 19.04\\
            \hline
            Pos. C & 2.28 & 1.97 &  12.31 \\
            \hline
            Other & 60.35 & 42.72 & 46.85\\
            \hline
        \end{tabular}}
        \captionof{subtable}{For participant 2}
        \label{tab:gaze_b2}
    \end{minipage}
    \vspace{2pt}
    \begin{minipage}{0.4\textwidth}
        \centering
        \small
        \resizebox{\textwidth}{!}{
        \begin{tabular}{|c|c|c|c|}
            \hline
            & Solo & Handover & Joint \\
            \hline
            Robot & 3.60 &  18.04 & 6.71 \\
            \hline
            Pos. B &  51.55 & 21.33 & 13.19\\
            \hline
            Pos. C & 13.82 & 31.36 &  51.89 \\
            \hline
            Other & 31.03 & 29.27 & 28.21\\
            \hline
        \end{tabular}}
        \captionof{subtable}{For participant 3}
        \label{tab:gaze_b3}
    \end{minipage}
    \hspace{35pt}
    \begin{minipage}{0.4\textwidth}
        \centering
        \small
        \resizebox{\textwidth}{!}{
        \begin{tabular}{|c|c|c|c|}
            \hline
            & Solo & Handover & Joint \\
            \hline
            Robot & 10.13 &  24.95 & 21.81 \\
            \hline
            Pos. B &  21.36 & 23.45 & 13.42\\
            \hline
            Pos. C & 7.75 & 8.96 &  23.62 \\
            \hline
            Other & 60.75 & 42.64 & 53.14\\
            \hline
        \end{tabular}}
        \captionof{subtable}{For participant 4}
        \label{tab:gaze_b4}
    \end{minipage}
    \vspace{2pt}
    \begin{minipage}{0.4\textwidth}
        \centering
        \small
        \resizebox{\textwidth}{!}{
        \begin{tabular}{|c|c|c|c|}
            \hline
            & Solo & Handover & Joint \\
            \hline
            Robot & 3.40 &  12.49 & 5.07 \\
            \hline
            Pos. B &  4.05 & 12.50 & 3.79\\
            \hline
            Pos. C & 5.07 & 3.05 &  5.15 \\
            \hline
            Other & 87.50 & 71.96 & 85.99\\
            \hline
        \end{tabular}}
        \captionof{subtable}{For participant 5}
        \label{tab:gaze_b5}
    \end{minipage}
    \hspace{35pt}
    \begin{minipage}{0.4\textwidth}
        \centering
        \small
        \resizebox{\textwidth}{!}{
        \begin{tabular}{|c|c|c|c|}
            \hline
            & Solo & Handover & Joint \\
            \hline
            Robot & 32.06 &  52.94 & 36.18 \\
            \hline
            Pos. B & 19.83 & 13.93 & 16.42\\
            \hline
            Pos. C & 1.19 & 5.28 &  15.38 \\
            \hline
            Other & 46.20 & 27.85 & 32.01\\
            \hline
        \end{tabular}}
        \captionof{subtable}{For participant 6}
        \label{tab:gaze_b6}
    \end{minipage}
    \begin{minipage}{0.4\textwidth}
        \centering
        \small
        \resizebox{\textwidth}{!}{
        \begin{tabular}{|c|c|c|c|}
            \hline
            & Solo & Handover & Joint \\
            \hline
            Robot & 14.89 &  32.39 & 14.12 \\
            \hline
            Pos. B &  17.84 & 16.76 & 10.98\\
            \hline
            Pos. C & 1.10 & 4.23 &  14.97 \\
            \hline
            Other & 66.17 & 46.62 & 59.94\\
            \hline
        \end{tabular}}
        \captionof{subtable}{For participant 7}
        \label{tab:gaze_b7}
    \end{minipage}
    \hspace{35pt}
    \begin{minipage}{0.4\textwidth}
        \centering
        \small
        \resizebox{\textwidth}{!}{
        \begin{tabular}{|c|c|c|c|}
            \hline
            & Solo & Handover & Joint \\
            \hline
            Robot & 29.28 &  58.16 & 30.16 \\
            \hline
            Pos. B & 34.02 & 11.80 & 16.27\\
            \hline
            Pos. C & 1.16 & 1.97 & 19.37 \\
            \hline
            Other & 35.60 & 28.06 & 34.20\\
            \hline
        \end{tabular}}
        \captionof{subtable}{For participant 8}
        \label{tab:gaze_b8}
    \end{minipage}
        \begin{minipage}{0.4\textwidth}
        \centering
        \small
        \resizebox{\textwidth}{!}{
        \begin{tabular}{|c|c|c|c|}
            \hline
            & Solo & Handover & Joint \\
            \hline
            Robot & 14.89 & 32.39 & 14.12 \\
            \hline
            Pos. B & 17.84 & 16.76 & 10.98\\
            \hline
            Pos. C & 1.10 & 4.23 & 14.97 \\
            \hline
            Other & 66.17 & 46.62 & 59.94\\
            \hline
        \end{tabular}}
        \captionof{subtable}{For participant 9}
        \label{tab:gaze_b7}
    \end{minipage}
    \hspace{35pt}
    \begin{minipage}{0.4\textwidth}
        \centering
        \small
        \resizebox{\textwidth}{!}{
        \begin{tabular}{|c|c|c|c|}
            \hline
            & Solo & Handover & Joint \\
            \hline
            Robot & 29.28 & 58.16 & 30.16 \\
            \hline
            Pos. B & 34.02 & 11.80 & 16.27\\
            \hline
            Pos. C & 1.16 & 1.97 & 19.37 \\
            \hline
            Other & 35.60 & 28.06 & 34.20\\
            \hline
        \end{tabular}}
        \captionof{subtable}{For participant 10}
        \label{tab:gaze_b8}
    \end{minipage}
        \begin{minipage}{0.4\textwidth}
        \centering
        \small
        \resizebox{\textwidth}{!}{
        \begin{tabular}{|c|c|c|c|}
            \hline
            & Solo & Handover & Joint \\
            \hline
            Robot & 33.23 & 53.84 & 23.03 \\
            \hline
            Pos. B & 23.17 & 4.64 & 8.18\\
            \hline
            Pos. C & 5.20 & 1.35 & 18.81 \\
            \hline
            Other & 38.41 & 40.17 & 49.98\\
            \hline
        \end{tabular}}
        \captionof{subtable}{For participant 11}
        \label{tab:gaze_b7}
    \end{minipage}
    \hspace{35pt}
    \begin{minipage}{0.4\textwidth}
        \centering
        \small
        \resizebox{\textwidth}{!}{
        \begin{tabular}{|c|c|c|c|}
            \hline
            & Solo & Handover & Joint \\
            \hline
            Robot & 20.02 & 43.66 & 20.68 \\
            \hline
            Pos. B & 35.36 & 18.63 & 30.12\\
            \hline
            Pos. C & 2.57 & 1.85 & 14.53 \\
            \hline
            Other & 42.05 & 35.85 & 34.67\\
            \hline
        \end{tabular}}
        \captionof{subtable}{For participant 12}
        \label{tab:gaze_b8}
    \end{minipage}
    \begin{minipage}{0.4\textwidth}
        \centering
        \small
        \resizebox{\textwidth}{!}{
        \begin{tabular}{|c|c|c|c|}
            \hline
            & Solo & Handover & Joint \\
            \hline
            Robot & 9.47 & 34.34 & 19.20 \\
            \hline
            Pos. B & 31.91 & 7.75 & 10.20\\
            \hline
            Pos. C & 3.58 & 7.96 & 16.34 \\
            \hline
            Other & 55.03 & 49.94 & 54.25\\
            \hline
        \end{tabular}}
        \captionof{subtable}{For participant 14}
        \label{tab:gaze_b7}
    \end{minipage}
    \hspace{35pt}
    \begin{minipage}{0.4\textwidth}
        \centering
        \small
        \resizebox{\textwidth}{!}{
        \begin{tabular}{|c|c|c|c|}
            \hline
            & Solo & Handover & Joint \\
            \hline
            Robot & 23.01 & 23.64 & 12.60 \\
            \hline
            Pos. B & 17.84 & 21.30 & 10.05\\
            \hline
            Pos. C & 2.23 & 6.90 & 24.28 \\
            \hline
            Other & 56.92 & 48.16 & 53.07\\
            \hline
        \end{tabular}}
        \captionof{subtable}{For participant 15}
        \label{tab:gaze_b8}
    \end{minipage}
\end{table}

\begin{table}[H]
    \centering
    \caption{Percentages of Gaze, Time t=-5 to 0 s, before movement onset}
    \label{tab:participant_all_gaze_before}
    \begin{minipage}{0.4\textwidth}
        \centering
        \resizebox{\textwidth}{!}{
        \begin{tabular}{|c|c|c|c|}
            \hline
            & Solo & Handover & Joint \\
            \hline
            Robot & 6.74 & 15.68 & 3.90 \\
            \hline
            Pos. B & 9.17 & 13.97 & 6.80 \\
            \hline
            Pos. C & 0.35 & 0.67 & 11.15 \\
            \hline
            Other & 83.74 & 69.69 & 78.16 \\
            \hline
        \end{tabular}}
        \captionof{subtable}{For participant 1}
        \label{tab:gaze_a1}
    \end{minipage}
    \hspace{35pt}
    \begin{minipage}{0.4\textwidth}
        \centering
        \resizebox{\textwidth}{!}{
        \begin{tabular}{|c|c|c|c|}
            \hline
            & Solo & Handover & Joint \\
            \hline
            Robot       & 4.70 & 12.93 & 4.20 \\
            \hline
            Pos. B       & 22.22 & 5.36 & 4.72 \\
            \hline
            Pos. C       & 3.06 & 3.88 & 17.21 \\
            \hline
            Other       & 70.02 & 77.83 & 73.87 \\
            \hline
        \end{tabular}}
        \captionof{subtable}{For participant 2}
        \label{tab:gaze_a2}
    \end{minipage}
    \vspace{2pt}
    \begin{minipage}{0.4\textwidth}
        \centering
        \resizebox{\textwidth}{!}{
        \begin{tabular}{|c|c|c|c|}
            \hline
            & Solo & Handover & Joint \\
            Robot       & 1.68 & 22.95 & 1.73 \\
            \hline
            Pos. B       & 47.56 & 5.80 & 3.20 \\
            \hline
            Pos. C       & 9.34 & 19.70 & 49.69 \\
            \hline
            Other       & 41.42 & 51.55 & 45.38 \\
            \hline
        \end{tabular}}
        \captionof{subtable}{For participant 3}
        \label{tab:gaze_a3}
    \end{minipage}
    \hspace{35pt}
    \begin{minipage}{0.4\textwidth}
        \centering
        \resizebox{\textwidth}{!}{
        \begin{tabular}{|c|c|c|c|}
            \hline
            & Solo & Handover & Joint \\
            \hline
            Robot       & 7.47 & 12.21 & 3.49 \\
            \hline
            Pos. B       & 9.25 & 9.70 & 6.35 \\
            \hline
            Pos. C       & 3.04 & 5.12 & 7.56 \\
            \hline
            Other       & 80.24 & 72.97 & 82.60 \\
            \hline
        \end{tabular}}
        \captionof{subtable}{For participant 4}
        \label{tab:gaze_a4}
    \end{minipage}
    \vspace{2pt}
    \begin{minipage}{0.4\textwidth}
        \centering
        \resizebox{\textwidth}{!}{
        \begin{tabular}{|c|c|c|c|}
            \hline
            & Solo & Handover & Joint \\
            \hline
            Robot       & 0.00 & 0.03 & 0.00 \\
            \hline
            Pos. B       & 0.00 & 0.00 & 0.00 \\
            \hline
            Pos. C       & 0.00 & 0.00 & 0.00 \\
            \hline
            Other       & 100.00 & 99.97 & 100.00 \\
            \hline
        \end{tabular}}
        \captionof{subtable}{For participant 5}
        \label{tab:gaze_a5}
    \end{minipage}
    \hspace{35pt}
    \begin{minipage}{0.4\textwidth}
        \centering
        \resizebox{\textwidth}{!}{
        \begin{tabular}{|c|c|c|c|}
            \hline
            & Solo & Handover & Joint \\
            \hline
            Robot       & 20.30 & 37.03 & 16.15 \\
            \hline
            Pos. B       & 26.00 & 14.88 & 14.59 \\
            \hline
            Pos. C       & 1.05 & 2.07 & 20.41 \\
            \hline
            Other       & 52.65 & 46.01 & 48.85 \\
            \hline
        \end{tabular}}
        \captionof{subtable}{For participant 6}
        \label{tab:gaze_a6}
    \end{minipage}
    \begin{minipage}{0.4\textwidth}
        \centering
        \resizebox{\textwidth}{!}{
        \begin{tabular}{|c|c|c|c|}
            \hline
            & Solo & Handover & Joint \\
            \hline
            Robot       & 1.36 & 10.89 & 0.79 \\
            \hline
            Pos. B       & 2.84 & 0.39 & 0.76 \\
            \hline
            Pos. C       & 0.10 & 0.58 & 1.63 \\
            \hline
            Other       & 95.70 & 88.14 & 96.82 \\
            \hline
        \end{tabular}}
        \captionof{subtable}{For participant 7}
        \label{tab:gaze_a7}
    \end{minipage}
    \hspace{35pt}
    \begin{minipage}{0.4\textwidth}
        \centering
        \resizebox{\textwidth}{!}{
        \begin{tabular}{|c|c|c|c|}
            \hline
            & Solo & Handover & Joint \\
            \hline
            Robot       & 11.85 & 15.28 & 2.10 \\
            \hline
            Pos. B       & 17.36 & 4.25 & 2.39 \\
            \hline
            Pos. C       & 2.28 & 6.40 & 18.74 \\
            \hline
            Other       & 68.50 & 74.07 & 76.77 \\
            \hline
        \end{tabular}}
        \captionof{subtable}{For participant 8}
        \label{tab:gaze_a8}
    \end{minipage}
        \begin{minipage}{0.4\textwidth}
        \centering
        \resizebox{\textwidth}{!}{
        \begin{tabular}{|c|c|c|c|}
            \hline
            & Solo & Handover & Joint \\
            \hline
            Robot       & 19.22 & 30.81 & 36.90 \\
            \hline
            Pos. B       & 21.78 & 4.25 & 2.38 \\
            \hline
            Pos. C       & 0.00 & 0.08 & 5.26 \\
            \hline
            Other       & 59.00 & 64.86 & 55.46 \\
            \hline
        \end{tabular}}
        \captionof{subtable}{For participant 9}
        \label{tab:gaze_a9}
    \end{minipage}
    \hspace{35pt}
    \begin{minipage}{0.4\textwidth}
        \centering
        \resizebox{\textwidth}{!}{
        \begin{tabular}{|c|c|c|c|}
            \hline
            & Solo & Handover & Joint \\
            \hline
            Robot       & 16.08 & 42.57 & 12.15 \\
            \hline
            Pos. B       & 26.08 & 4.38 & 10.18 \\
            \hline
            Pos. C       & 0.79 & 1.21 & 17.22 \\
            \hline
            Other       & 57.05 & 51.84 & 60.45 \\
            \hline
        \end{tabular}}
        \captionof{subtable}{For participant 10}
        \label{tab:gaze_a10}
    \end{minipage}
        \begin{minipage}{0.4\textwidth}
        \centering
        \resizebox{\textwidth}{!}{
        \begin{tabular}{|c|c|c|c|}
            \hline
            & Solo & Handover & Joint \\
            \hline
            Robot       & 15.83 & 31.23 & 9.00 \\
            \hline
            Pos. B       & 16.50 & 2.02 & 6.14 \\
            \hline
            Pos. C       & 1.00 & 0.81 & 14.15 \\
            \hline
            Other       & 66.67 & 65.93 & 70.71 \\
            \hline
        \end{tabular}}
        \captionof{subtable}{For participant 11}
        \label{tab:gaze_a11}
    \end{minipage}
    \hspace{35pt}
    \begin{minipage}{0.4\textwidth}
        \centering
        \resizebox{\textwidth}{!}{
        \begin{tabular}{|c|c|c|c|}
            \hline
            & Solo & Handover & Joint \\
            \hline
            Robot       & 20.55 & 26.67 & 20.34 \\
            \hline
            Pos. B       & 8.35 & 5.96 & 7.33 \\
            \hline
            Pos. C       & 1.18 & 1.44 & 11.39 \\
            \hline
            Other       & 69.92 & 65.93 & 60.94 \\
            \hline
        \end{tabular}}
        \captionof{subtable}{For participant 12}
        \label{tab:gaze_a12}
    \end{minipage}
    \begin{minipage}{0.4\textwidth}
        \centering
        \resizebox{\textwidth}{!}{
        \begin{tabular}{|c|c|c|c|}
            \hline
            & Solo & Handover & Joint \\
            \hline
            Robot       & 12.99 & 2.81 & 3.12 \\
            \hline
            Pos. B       & 2.60 & 1.31 & 1.94 \\
            \hline
            Pos. C       & 1.08 & 0.97 & 4.78 \\
            \hline
            Other       & 83.33 & 94.91 & 90.16 \\
            \hline
        \end{tabular}}
        \captionof{subtable}{For participant 14}
        \label{tab:gaze_a14}
    \end{minipage}
    \hspace{35pt}
    \begin{minipage}{0.4\textwidth}
        \centering
        \resizebox{\textwidth}{!}{
        \begin{tabular}{|c|c|c|c|}
            \hline
            & Solo & Handover & Joint \\
            \hline
            Robot       & 6.94 & 5.57 & 2.81 \\
            \hline
            Pos. B       & 10.32 & 3.38 & 4.98 \\
            \hline
            Pos. C       & 2.77 & 2.80 & 12.09 \\
            \hline
            Other       & 79.97 & 88.25 & 80.12 \\
            \hline
        \end{tabular}}
        \captionof{subtable}{For participant 15}
        \label{tab:gaze_a15}
    \end{minipage}
\end{table}

\subsection*{}
\textbf{Statistical Analysis of EEG Frequency Bands}
\begin{table}[h!]

    \centering
    \caption{Average ERDS of mu, beta, and gamma bands in the time interval [-2, 0] seconds for two conditions: Handover and Non-Handover}
    \resizebox{0.70\columnwidth}{!}{
    \begin{tabular}{ccccc}
    \toprule
        SubjectID & Condition & mu & beta & gamma \\
        \midrule
        \multirow{2}{*}{S1} & Handover & 0.083 & 0.179 & 0.116 \\& NonHandover &  0.161 & 0.218 & 0.166 \\
        \midrule
        \multirow{2}{*}{S2} & Handover & 0.138 & 0.133 & 0.051 \\& NonHandover & 0.138 & 0.233 & 0.105 \\
        \midrule
        \multirow{2}{*}{S3} & Handover & 0.199 & 0.286 & 0.321 \\& NonHandover & 0.173 & 0.195 & 0.173 \\
        \midrule
        \multirow{2}{*}{S4} & Handover & 0.123 & 0.234 & 0.156 \\& NonHandover & 0.190 & 0.240 & 0.173 \\
        \midrule
        \multirow{2}{*}{S5} & Handover & 0.194 & 0.187 & 0.214 \\& NonHandover & 0.100 & 0.233 & 0.164 \\
        \midrule
        \multirow{2}{*}{S6} & Handover & 0.163 & 0.092 & 0.147 \\& NonHandover & 0.134 & 0.116 & 0.099 \\
        \midrule
        \multirow{2}{*}{S8} & Handover & 0.141 & 0.099 & 0.159 \\& NonHandover & 0.094 & 0.126 & 0.122 \\
        \midrule
        \multirow{2}{*}{S9} & Handover & 0.058 & 0.418 & 0.323 \\& NonHandover & 0.074 & 0.201 & 0.124 \\
        \midrule
        \multirow{2}{*}{S10} & Handover & 0.680 & 0.265 & 0.344 \\& NonHandover & 0.236 & 0.185 & 0.176 \\
        \midrule
        \multirow{2}{*}{S11} & Handover & 0.072 & 0.252 & 0.132 \\& NonHandover & 0.184 & 0.222 & 0.134 \\
        \midrule
        \multirow{2}{*}{S12} & Handover & 0.100 & 0.155 & 0.189 \\& NonHandover & 0.173 & 0.254 & 0.451 \\
        \midrule
        \multirow{2}{*}{S14} & Handover & 0.102 & 0.097 & 0.110 \\& NonHandover & 0.140 & 0.061 & 0.154 \\
        \midrule
        \multirow{2}{*}{S15} & Handover & 0.240 & 0.184 & 0.173 \\& NonHandover & 0.121 & 0.163 & 0.149 \\
        \bottomrule
    \end{tabular}
    }
    \label{tab:erds}
\end{table}

\newpage
\subsection*{Supplementary Figures}

\subsection*{}
\textbf{Different Human Actions Included in the Experiment}
\begin{figure}[H]
       \centering
         \includegraphics[width=.8\linewidth,trim={1.5cm 1.5cm 1.75cm 1.75cm},clip]{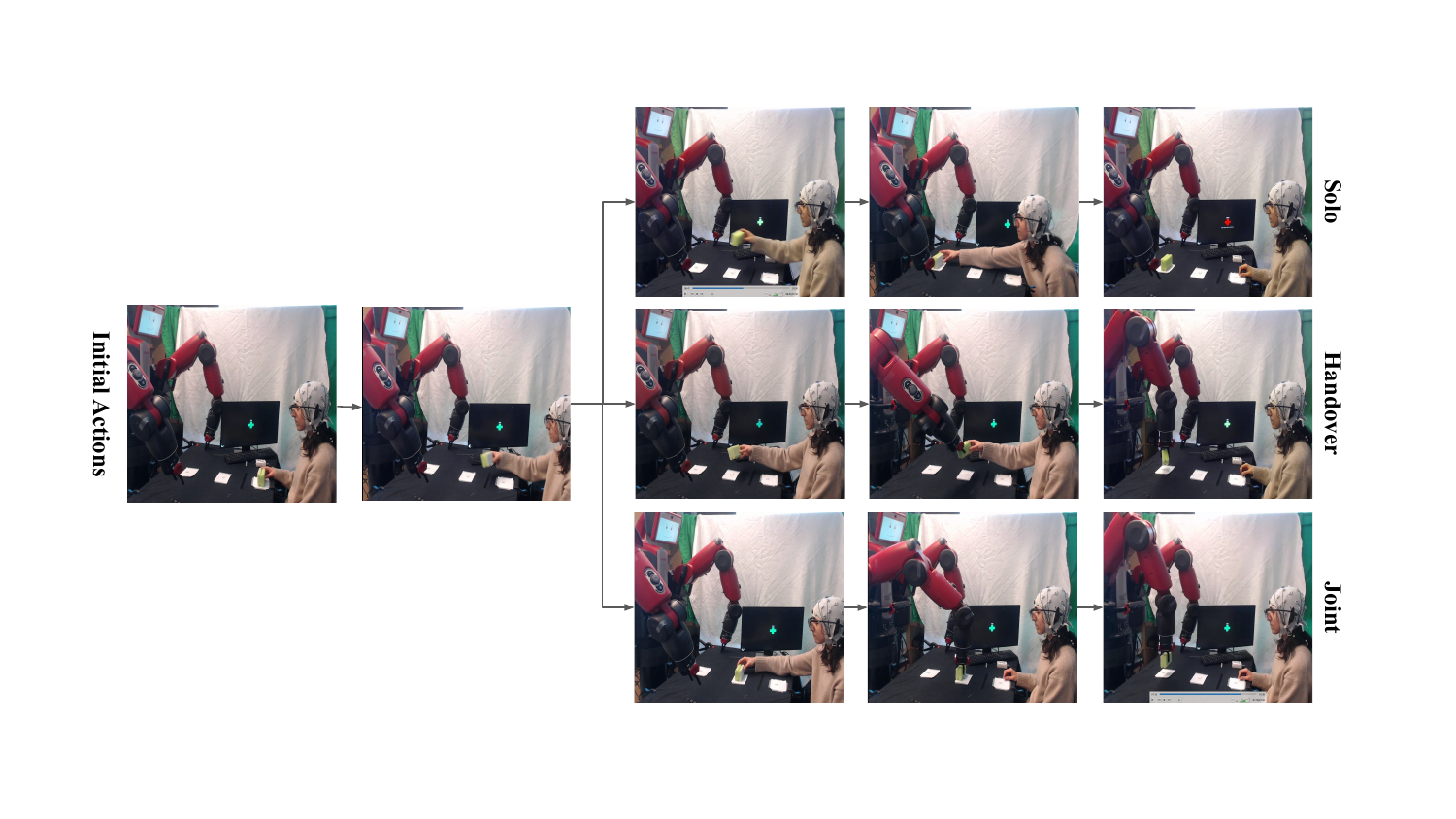}  
     \caption{Overview of different actions in our experiment. After the \emph{Go} signal, the user picks up the object. The initial action is the same for all cases. In \textbf{Solo action}, the user carries the object and places it at B. In \textbf{Handover}, the user carries the object to C, where the Robotic arm approaches. The user hands it to the Robot, which then places it at B. In \textbf{Joint action}, the user places the object at C, and the Robot picks it up from C and places it at B (Fig. 2 in (\emph{14}))}
    \label{fig:experiment_overview}
\end{figure}

\subsection*{}
\textbf{The Location of EEG Electrodes on the Scalp}
\begin{figure}[H]
    \centering
    \includegraphics[height=0.37\linewidth]{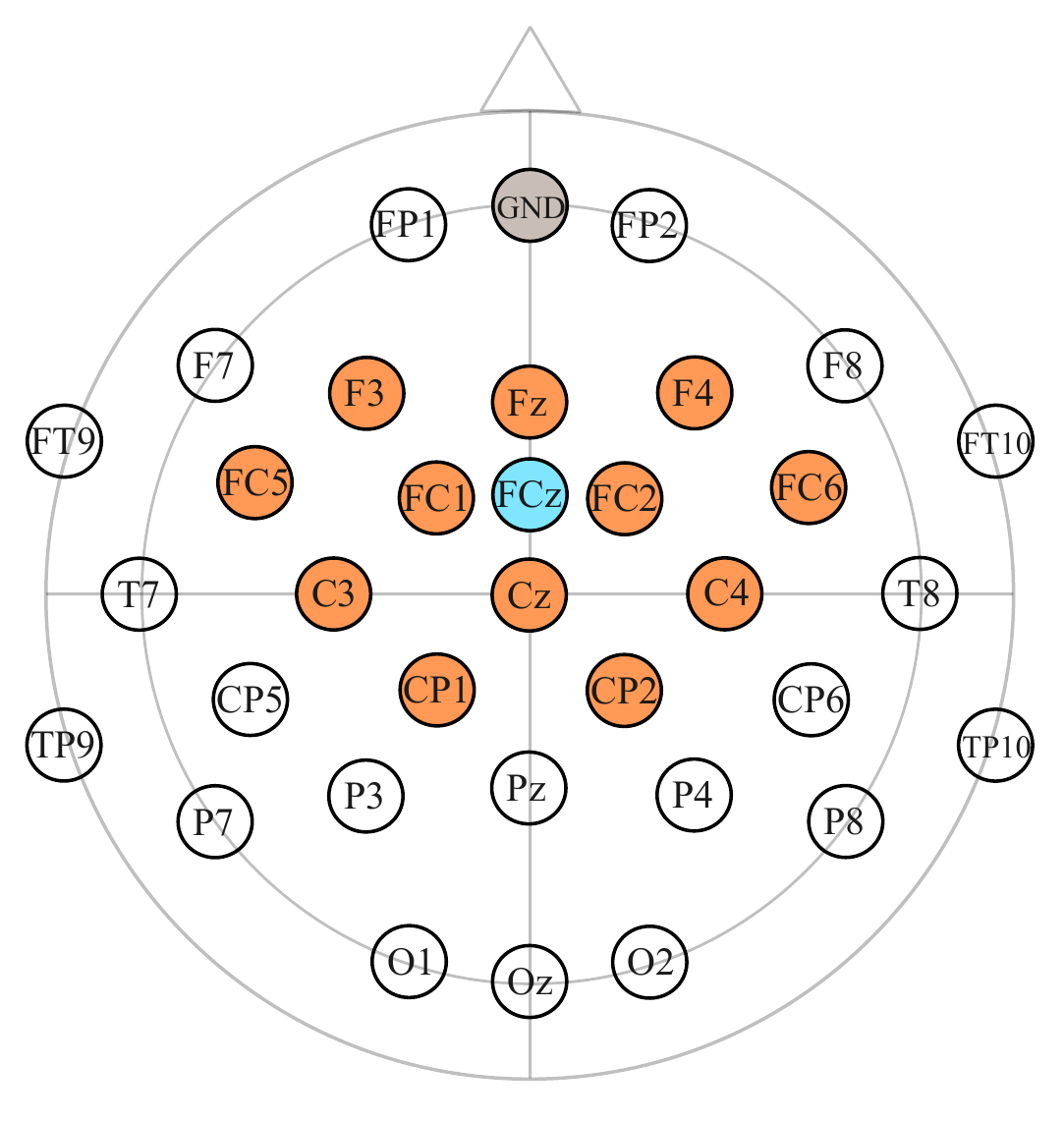}
    \caption{EEG electrodes locations on the scalp. The orange color indicates the electrodes that were used in the analysis and all other electrodes were excluded.}
    \label{fig:eeg_channels}
\end{figure}

\subsection*{}
\textbf{First Person View Point Captured by the Eye-Tracker Glasses}
\begin{figure}[H]
    \centering
    \includegraphics[width=0.75\linewidth]{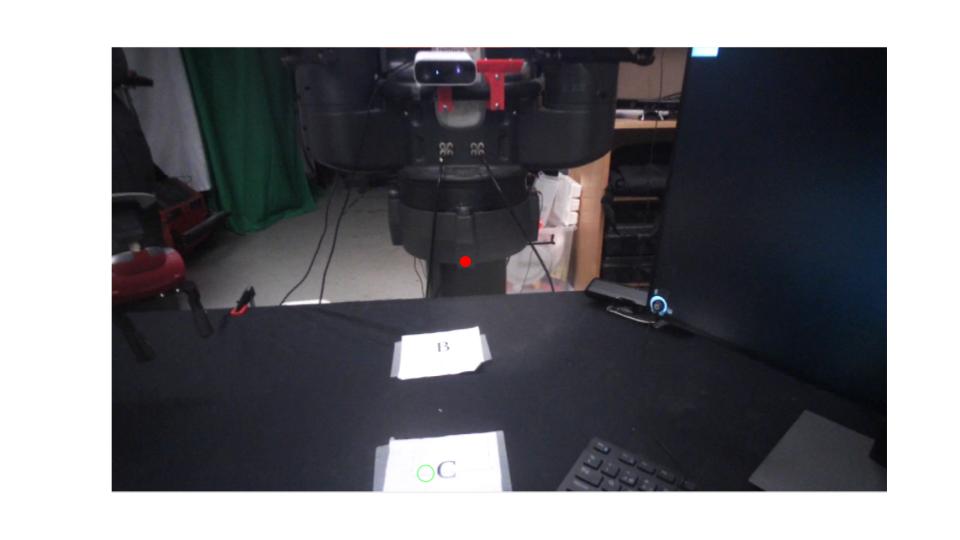}
    \caption{First-person view of a participant performing the experiment. The green circle (on position C) represents the tracked gaze, while the red dot on the torso of the Baxter Robot indicates the reference point for the gaze data.}
    \label{fig:gaze_tracking_POV}
\end{figure}

\end{document}